\documentclass[conference]{IEEEtran}
\usepackage{times}

\usepackage{amsmath}
\usepackage{amssymb}
\usepackage{mathtools}

\usepackage{tabularx}
\usepackage{graphicx}
\usepackage{bm}
\usepackage{color}
\usepackage{float}
\usepackage{units}
\usepackage{url}
\usepackage{hyperref}
\usepackage{balance}
\usepackage{amsmath}
\usepackage[numbers]{natbib}
\usepackage{algorithm}
\usepackage{algpseudocode}
\usepackage{cancel}
\usepackage{wasysym}
\usepackage{makecell}

\makeatletter 
\def\endfigure{\end@float} 
\def\endtable{\end@float}
\makeatother 
\newtheorem{assumption}{Assumption}

\usepackage{subcaption}
\usepackage[]{graphicx}
\usepackage{caption}
\renewcommand{\unit}[1]{{\rm #1} }

\newtheorem{remark}{\textbf{Remark}}

\begin{document} 

\title{			
  Gait-Net-augmented Implicit Kino-dynamic MPC \\ for Dynamic Variable-frequency Humanoid Locomotion over Discrete Terrains
}

\author{Junheng Li$^\dagger$, Ziwei Duan, Junchao Ma, and Quan Nguyen \vspace{0.1cm} \\University of Southern California, USA \\
$^\dagger$Corresponding Author.\quad  Emails: {\tt\small \{junhengl,ziweidua,junchaom,quann\}@usc.edu}.}

\maketitle

\begin{abstract}

Reduced-order-model-based optimal control techniques for humanoid locomotion struggle to adapt step duration and placement simultaneously in dynamic walking gaits due to their reliance on fixed-time discretization, which limits responsiveness to various disturbances and results in suboptimal performance in challenging conditions.
In this work, we propose a Gait-Net-augmented implicit kino-dynamic model-predictive control (MPC) to simultaneously optimize step location, step duration, and contact forces for natural variable-frequency locomotion.
The proposed method incorporates a Gait-Net-augmented Sequential Convex MPC algorithm to solve multi-linearly constrained variables by iterative quadratic programs. At its core, a lightweight Gait-frequency Network (Gait-Net) determines the preferred step duration in terms of variable MPC sampling times, simplifying step duration optimization to the parameter level. 
Additionally, it enhances and updates the spatial reference trajectory within each sequential iteration by incorporating local solutions, allowing the projection of kinematic constraints to the design of reference trajectories. 
We validate the proposed algorithm in high-fidelity simulations and on small-size humanoid hardware, demonstrating its capability for variable-frequency and 3-D discrete terrain locomotion with only a one-step preview of terrain data.

\end{abstract}


\section{Introduction}
\label{sec:Introduction}


Contact/footstep planning is a fundamental problem in humanoid robot motion control. Due to the inherent instability of these robots, 
efficient integration of both contact planning and motion control is an essential step to enable dynamic and robust locomotion across diverse terrains.

\subsection{Contact Planning}
Footstep planning has traditionally been treated as a high-level, standalone problem solved before motion control execution \cite{deits2014footstep, bouyarmane2012humanoid, carpentier2016versatile, ponton2021efficient}. For instance, \citet{deits2014footstep} addresses 3D humanoid footstep planning using a highly efficient mixed-integer quadratically constrained quadratic program (MIQCQP) to compute trajectories in seconds. In addition, \citet{bouyarmane2012humanoid} proposes a best-first algorithm for collision-free multi-contact planning in humanoid loco-manipulation. However, these offline trajectories, tracked by whole-body control or inverse-kinematics-based controllers, often suffer from error accumulation over long trajectories due to their open-loop nature.

Contact-implicit model-predictive control (CI-MPC) has gained popularity for optimizing contact force, location, and timing within a unified framework by addressing the linear complementarity problem (LCP) between contact velocity and force \cite{kim2023contact, le2024fast, kong2023hybrid}. These frameworks integrate contact planning and motion control to generate optimal contact behaviors in real-time. However, CI-MPC involves solving highly complex, nonlinear problems with significant computational demands. While \citet{le2024fast} precompute LCP parameters offline to improve online speed, this process remains costly and must be repeated for each new setup.

\begin{figure}[!t]
\vspace{0cm}
    \center
    \includegraphics[clip, trim=0cm 4cm 17.5cm 0.1cm, width=1\columnwidth]{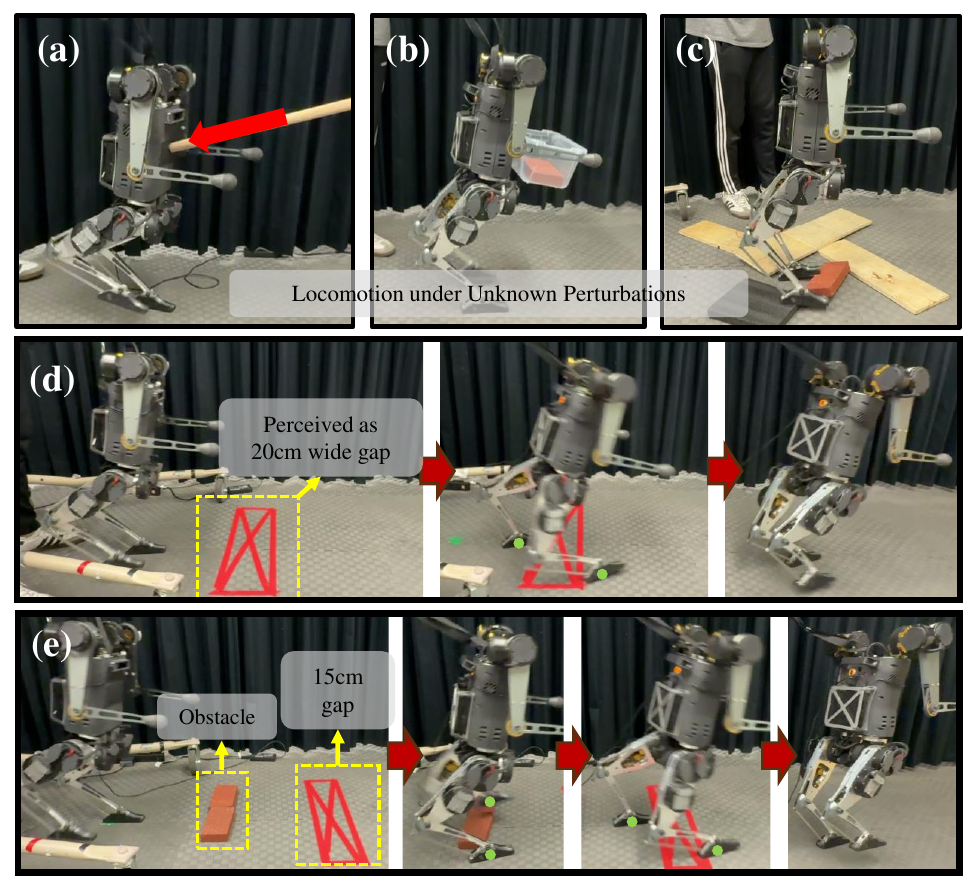}
    \caption{{\bfseries Gait-Net-augmented Kino-dynamic MPC.} Hardware Experiment Snapshots. (a). Push-recovery in locomotion (b). Locomotion while carrying an unknown 0.75 kg object; (c). Locomotion over unknown uneven terrain; (d). Dynamic walking over terrain with a 20 cm terrain gap at 0.75 m/s; (e). Dynamic walking over terrain gap and obstacle. Full experiment video: \url{https://youtu.be/UqLDYHGL5EA} }
    \label{fig:title}
    \vspace{-0.2cm}
\end{figure}

\subsection{Humanoid Locomotion with Simplified Models}
Convex MPC (CMPC) is widely used in legged locomotion control, leveraging linearized dynamics and constraints, such as single rigid-body model (SRBM) \cite{di2018dynamic}, for fast and high-frequency control by optimizing ground reaction forces. However, due to the bilinear coupling of force and foot location vectors, predefined contact locations are required \cite{di2018dynamic, ding2022orientation} for linearization. Foot placement is typically determined separately using heuristics (\textit{e.g.}, Raibert heuristic \cite{raibert1986legged}), capture point methods \cite{pratt2006capture}, or optimizations with the linear inverted pendulum model \cite{gu2024walking}. These approaches embed predefined step durations (\textit{a.k.a.}, one-step gait duration), making it challenging to invert the problem and determine duration from foot location, especially in 3-D. Feed-forward step duration is also suboptimal for adaptive strategies on uneven terrain, where step duration should highly correlate with stride length and current foot actuation.

Centroidal dynamics (CD) is widely used in humanoid robot control for its simplified yet effective representation of whole-body dynamics, making it suitable for real-time planning and control \cite{orin2013centroidal}. CD-MPC and kino-dynamic MPC are often formulated as nonlinear MPC (NMPC) problems \cite{romualdi2022online, dai2014whole, elobaid2023online}. \citet{garcia2021mpc} leverages CMPC formulation and CD-augmented SRBM to include link inertia.  However, the framework still requires predefined gait schedules and the generation of footstep locations from separate modules. Additionally, Kino-dynamic MPC offers an advantage over CD-MPC by explicitly optimizing joint states, ensuring feasible whole-body motions \cite{dantec2024centroidal}. In contrast, CD-MPC lacks kinematic coupling between foot location and floating base states, often necessitating lower-level inverse kinematics (IK) motion generation \cite{meduri2023biconmp} or whole-body control \cite{wensing2016improved}.
In our work, the implicit kino-dynamic MPC aids kinematic assurance and eliminates the need to optimize joint states.

\begin{figure*}[!t]
\vspace{0.2cm}
		\center
		\includegraphics[clip, trim=0.5cm 0.3cm 0.5cm 0cm, width=1.8\columnwidth]{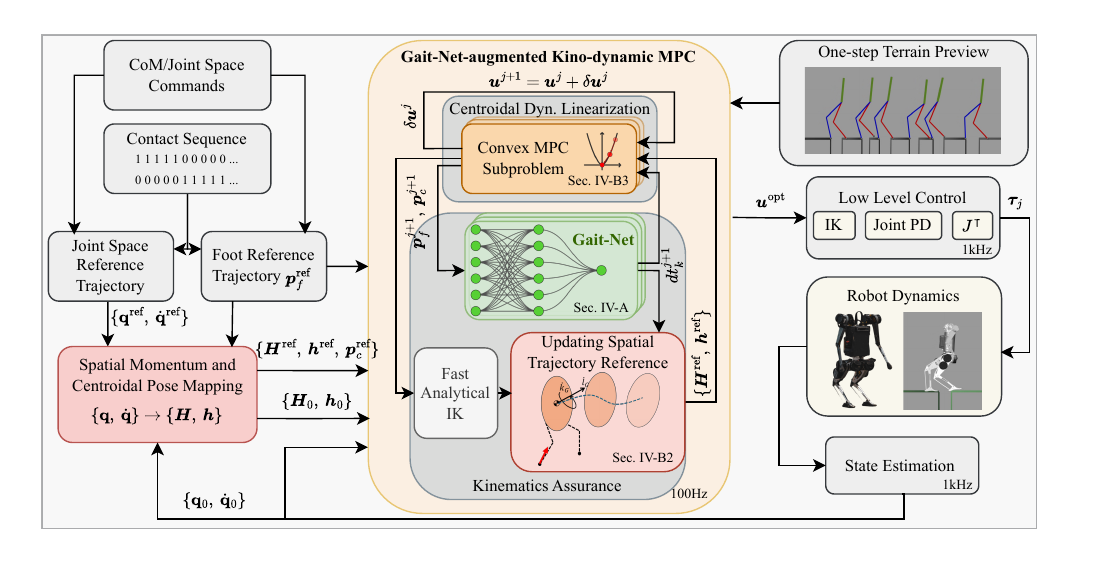}
		\caption{{\bfseries Control System Architecture}}
		\label{fig:controlArchi}
		\vspace{-0.2cm}
\end{figure*}

\subsection{Challenges in Variable Frequency Locomotion}
Variable gait frequency locomotion remains under-explored in humanoid robotics. 
Previous works have primarily addressed this problem at the footstep planning level \cite{khadiv2016step, griffin2017walking, nguyen2017dynamic, xiang2024adaptive}, optimizing only foot placement and timing.
While the MIT humanoid robot demonstrates impressive performance with whole-body MPC \cite{khazoom2024tailoring}, its gait frequency is fixed and determined by the MPC sampling time. \citet{li2023dynamic} optimize gait frequency offline for stepping stone terrain, but the open-loop approach lacks real-time adjustments based on state feedback. In contrast, our work enables variable-frequency bipedal walking by concurrently optimizing foot location, contact force, and gait frequency compactly and efficiently, addressing the motion control and variable foot-step planning together in one optimization.

Gait frequency can be integrated into the MPC framework by optimizing the sampling time per step with a fixed contact schedule. However, this introduces a multi-linear coupling of foot location, contact force, and sampling time, significantly increasing problem complexity and solving time. To address this, we propose a customized Neural-network-augmented sequential QP (SQP) solver that efficiently handles these multi-linear terms through iteratively solving QP problems.

\subsection{Learning-augmented Optimization-based Control}

Learning-based control methods, such as reinforcement learning (RL), have achieved significant success in legged locomotion \cite{li2021reinforcement, margolis2024rapid, krishna2022linear, bao2024deep}. 
For example, \citet{wang2025beamdojo} demonstrated natural locomotion over discrete stepping-stone terrains. However, the terrain configurations were limited to segments encountered during the second stage of training.
On the other hand, model-based approaches have also gained from learning-based techniques \cite{bang2024variable, el2024real, chen2024learning, romualdi2024online}. For example, pre-trained neural networks (NN) can approximate complex, computationally intensive nonlinear functions, as shown in \cite{bang2024variable}, where an NN predicts centroidal inertia evolution, eliminating the need for complex spatial momentum computations in optimization. In this work, we address the weak correlation between gait frequency and foot location, an inherent limitation of simplified dynamics models such as SRBM and CD, by introducing Gait-Net, an NN that is trained on data from variable-gait-frequency MPC with whole-body dynamics.

\subsection{Contributions}
The main contributions are twofold. Firstly, we propose a novel Gait-Net-augmented Implicit Kino-dynamic MPC framework that concurrently optimizes foot contact force, foot location, and footstep duration. In this framework, we introduce supervised learning in a sequential CMPC algorithm to efficiently solve NMPC with multi-linear constraints. 

Secondly, we complement CD with a Gait-frequency Network (Gait-Net) to form an \textit{implicit} kino-dynamic MPC. In each sequential MPC iteration, Gait-Net (1) determines the preferred step duration in terms of MPC sampling time $dt$, transforming $dt$ from a decision variable to a parameter for efficiency; and (2) improves the estimation of reference spatial momentum and pose trajectories to mimic more closely as an explicit kino-dynamic approach while eliminating the joint angles as part of state variables.

Additionally, we validate the proposed approach through both simulation and hardware experiments on humanoid robots. Our controller demonstrates robustness against unknown disturbances, successfully handling uneven terrain, push recovery, and unknown loads. It enables the robot to push a 35 kg cart and traverse discrete terrains with gaps (discontinuities) up to 20 cm. 

The rest of the paper is organized as follows. Sec. \ref{sec:overview} presents the overview of the proposed control system architecture. Sec. \ref{sec:bg} outlines the background and preliminaries of the MPC-based locomotion control methods.  Sec. \ref{sec:approach} presents the main approaches, including the proposed kino-dynamics MPC, the Gait-Net, and the main algorithm of the Gait-Net-augmented sequential MPC. Sec. \ref{sec:Results} presents highlighted numerical and hardware validations.

\section{System Overview}
\label{sec:overview}

In this section, we present the control system architecture of the proposed framework, shown in Fig. \ref{fig:controlArchi}. 
Empirically, humanoid kino-dynamics MPC explicitly optimizes the joint states through kinematics constraints \cite{gu2025humanoid}, while traditional centroidal-dynamics MPC often requires subsequent inverse kinematics solver or whole-body control for motion execution. Both approaches employ nonlinear approaches to solve the optimization problem. In our framework, we proposed a Gait-Net-augmented sequential CMPC algorithm that translates the original nonlinear problem into convex sequential subproblems. With the additional assistance of Gait-Net, we reduce the optimization variable and mimic a natural step duration decision in each iteration. 

The control framework converts user commands and contact sequence into joint space references $\{\mathbf q_k^\text{ref} \in \mathbb R^{6+n_j},\: \dot{\mathbf q}_k^\text{ref} \in \mathbb R^{6+n_j}\}^h_{k = 0}$ and foot location reference trajectory $\{\bm p_f^\text{ref}\in \mathbb R^{3n_i}\}^h_{k = 0}$, where $n_j$ is the number of joints, $n_i$ is the number of contact/foot, and $h$ is a finite number of horizon. These joint-space trajectories, along with joint-space feedback states, are then translated into spatial momenta $\bm h\in \mathbb R^6$ and their primitive, the centroidal pose $\bm H\in \mathbb R^6$, which are the state variables used in the Gait-Net-augmented kino-dynamic MPC. Within the MPC, we break down the nonlinear dynamics constraints into sequential CMPC subproblems that can be solved through QP solvers. In each sequential iteration $j$, the Gait-Net predicts and updates the MPC sampling time $dt$ towards convergence and enables variable-frequency walking.
The spatial momentum and pose trajectories are updated at each iteration to reflect the kinematic configuration based on the iterative solution of $dt$, CoM location $\bm p_c \in \mathbb R^3$, and foot locations $\bm p_f\in \mathbb R^{3n_i}$,
providing a kinematically feasible reference. Once the terminal condition is met in the custom sequential solver, the control inputs are then mapped to motor commands in low-level control, which incorporates standard techniques such as inverse kinematics, contact Jacobian mapping, and joint-PD swing leg control \cite{di2018dynamic}. Notably, the full Gait-Net-augmented Kino-dynamic MPC is run at the beginning of each footstep to determine the step duration, the rest of the duration will incorporate the kino-dynamic MPC with the same MPC $dt$ throughout this very footstep.

\section{Preliminary}
\label{sec:bg}

In this section, we introduce the background of the whole-body MPC used during the variable-frequency walking data collection process and a general explicit kino-dynamics MPC formulation for humanoid robots.

\subsection{Whole-body Model Predictive Control}
\label{subsec:wbmpc}
Driven by the strong dynamics and kinematics correlation between the footstep location, duration, and whole-body coordination with Whole-body MPC on humanoid robots \cite{khazoom2024tailoring, dantec2024centroidal}, we leverage such a control paradigm to obtain high-fidelity humanoid locomotion results in simulation to serve as the training dataset for Gait-Net. We particularly focus on adapting this MPC to achieve variable MPC sampling times for each footstep (\textit{i.e.}, variable-frequency walking).
Notably, to achieve variable-frequency walking, we fix a periodic contact sequence of the locomotion to be every $h$ MPC time-steps. Therefore, we can allow variable step durations at each footstep by adjusting the sampling time $dt_k$ of every $h'=h/2$ time-steps, as a one-step gait duration.

The optimization variable $\mathbf X^\text{wb}$ includes the robot states in the generalized coordinates $\mathbf q \in \mathbb R^{6+n_j}$, their rates of change $\dot{\mathbf q} \in \mathbb R^{6+n_j}$, joint torque $\bm \tau_j \in \mathbb R^{n_j}$, and the constraint forces $\bm \lambda = \{\bm f_i\in \mathbb R^{3};\: \bm \tau_i\in \mathbb R^{3}\}^{n_i}_{i=0} $ (\textit{i.e.}, spatial contact wrench), 
\begin{align}
    \mathbf X^\text{wb} = \{ \mathbf q_k,\: \dot{\mathbf q}_k,\: \bm \tau_{j,k},\: \bm \lambda_k \}^{h}_{k=0}
\end{align}

The nonlinear optimization problem can be formulated as

\begin{alignat}{3}
\label{eq:NMPCcost}
\underset{\mathbf X^\text{wb}}{\text{min}} \: & \sum_{k = 0}^{h-1} \big\| \mathbf q_k-  \mathbf q^{\text{ref}}_k\big\|^2 _{\bm Q_1} + \big\| \bm{\tau}_{j,k}  \big\|^2 _{\bm Q_2} + \big\| \bm \lambda_k  \big\|^2 _{\bm Q_3} + \big\| \dot{\mathbf q}_k  \big\|^2 _{\bm Q_4}\\ 
    \nonumber
    \textrm{subj} & \textrm{ect to:}
\end{alignat}
\vspace{-0.35cm}
\begin{subequations}
\label{eq: NMPCconstraints}
\allowdisplaybreaks
\setlength\abovedisplayskip{-3pt}
\begin{alignat}{3}
    \label{eq:mpcDynamics}
    \textrm{Dynamics: } \quad & \{\dot{{\mathbf q}}_{k+1},\: {{\mathbf q}}_{k+1} \}= f^{\mathrm{wb}}(\mathbf X^\text{wb}_{k}, dt_k),\\
    \label{eq:mpcequality}
    \textrm{Eq. Cons.:} \quad & g^\text{wb}_\mathrm{eq}(\mathbf X^\text{wb}_k) = \bm b^\text{wb}_\mathrm{eq}, \\
    \label{eq:mpcinequality}
    \textrm{Ineq. Cons.:} \quad  & \bm b^\text{wb}_\mathrm{lb} \leq g^\text{wb}_\mathrm{ineq}(\mathbf X^\text{wb}_k) \leq \bm b^\text{wb}_\mathrm{ub}.
\end{alignat}
\end{subequations}

Equation (\ref{eq:NMPCcost}) refers to the objective function of the NMPC, which is to track a predefined trajectory, while minimizing joint torque, contact wrench, and rate of change of the states. Each objective is weighted by the corresponding diagonal matrix $\bm Q$. Equation (\ref{eq:mpcDynamics}) describes the discrete-time whole-body dynamics in the spatial vector form, modified from
\begin{align}
    \arraycolsep=1.4pt\def\arraystretch{1.25}
    \left[\begin{array}{cc} 
    \bm M(\mathbf q) & -\bm J_i^\intercal(\mathbf q) \\
    -\bm J_i^\intercal(\mathbf q) & \mathbf 0
    \end{array} \right]
    \left[\begin{array}{c} 
    \ddot{\mathbf q}\\
    \bm \lambda_i
    \end{array} \right] = 
    \left[\begin{array}{c} 
    -\bm C(\mathbf q, \dot{\mathbf q}) + \bm S \bm \tau_j \\
    \bm J_i(\mathbf q)\dot{\mathbf q}
    \end{array} \right] 
\end{align}
where $\bm M$ and $\bm C$ are the mass matrix and combination of the Coriolis terms and the gravity vector. $\bm J_i$ represents the spatial contact Jacobian of $i$th contact point. $\bm S$ is an actuation selection matrix in joint space.  

The optimization problem is also subjected to additional equality (\ref{eq:mpcequality}) and inequality (\ref{eq:mpcinequality}) constraints. These constraints include general constraints to ensure humanoid locomotion, such as friction pyramid constraints, torque limits, joint limits, contact wrench saturation, and contact wrench cone (CWC) constraints \cite{caron2015stability}. The joint torque result from optimization can be directly applied as the motor command to control the robot. Hence, the deployment frequency of such a controller has a direct impact on the control performance (\textit{i.e.}, a higher hyper-sample rate will benefit faster reaction to disturbances).

\subsection{Explicit Kino-dynamic Model Predictive Control}
\label{subsec:cdmpc}
Due to the heavy computation burden of the full-order dynamics model in whole-body MPC, many works that leverage this control method trade-off solution accuracy for high-frequency online deployment, such as constraining the number of iterations in SQP solvers \cite{khazoom2024tailoring, galliker2022planar} and leveraging Differential Dynamic Programming (DDP) \cite{dantec2022whole}. According to \cite{dantec2024centroidal}, the performance of such whole-body MPCs is still to be proven to outperform the relatively more computation-friendly Centroidal-dynamics MPC (CD-MPC) and Kino-dynamics MPC (full kinematics + CD) \cite{romualdi2022online, garcia2021mpc, he2024cdm, chignoli2021humanoid}. Hence, a well-constructed Kino-dynamics MPC can very well stand at the middle ground of computation intensity and control performance for online deployment.

The centroidal dynamics of a humanoid robot is:
\begin{align}
\label{eq:cd}
    \dot{\bm h}= \left[\begin{array}{c} 
    \dot{\bm l}_G\\
    \dot{\bm k}_G 
    \end{array} \right] = 
    \left[\begin{array}{c} 
    \sum_{i = 0}^{n_i}\bm f_i + m \bm g \\
    \sum_{i = 0}^{n_i}(\bm p_{f,i}-\bm p_c) \times \bm f_i + \bm \tau_i
    \end{array} \right],
\end{align}
where $\dot{\bm l}_G \in \mathbb R^3$ and $\dot{\bm k}_G \in \mathbb R^3$ are the rate of change of linear momentum and angular momentum. $(\bm p_{f,i}-\bm p_c)$ is the distance vector from the $i$th contact point $\bm p_{f,i}$ to robot CoM $\bm p_c$. 

Empirically, the \textit{explicit} kino-dynamics MPC requires joint states as part of the optimization variables, as the centroidal momentum matrix $\bm A_G$ \cite{orin2013centroidal} is dependent on the whole-body configuration, and
\begin{align}
\label{eq:hdot}
    \bm h  = \bm A_G(\mathbf q) \dot{\mathbf q}, \quad
    \dot {\bm h} =  \bm A_G(\mathbf q) \ddot{\mathbf q} + \dot{\bm A}_G(\mathbf q, \dot {\mathbf q}) \dot{\mathbf q}.
\end{align}

Hence the optimization variables of the explicit kino-dynamics MPC can be chosen as,
\begin{align}
    \mathbf X^\text{kd} = \{ \mathbf q,\: \dot{\mathbf q},\: \bm \lambda_k \}^{h}_{k=0},
\end{align}
 The finite horizon optimization problem with a prediction of $h$ steps can expressed as,
\begin{alignat}{3}
\label{eq:CDMPCcost}
\underset{\mathbf X^\text{kd}}{\text{min}} \: & \sum_{k = 0}^{h-1} 
\big\| \bm h_{k}^{\text{ref}} - \bm h_{k}  \big\|^2 _{\bm R_1}
+ \big\| \mathbf X_{k}^{\text{kd,ref}} - \mathbf X^\text{kd}_{k}  \big\|^2 _{\bm R_2} \\ 
\nonumber
\textrm{subject to:} & \quad
\end{alignat}
\vspace{-0.35cm}
\begin{subequations}
\label{eq: CDMPCconstraints}
\allowdisplaybreaks
\setlength\abovedisplayskip{-3pt}
\begin{alignat}{3}
    \label{eq:CDDynamics}
    \textrm{Dynamics: } \quad & \mathbf X^\text{kd}_{k+1} = f^{\mathrm{kd}}(\mathbf X^\text{kd}_{k}, dt_k),\\
    \label{eq:CDequality}
    \textrm{Eq. Cons.:} \quad & g^\mathrm{kd}_\mathrm{eq}(\mathbf X^\text{kd}_k) = \bm b^\mathrm{kd}_\mathrm{eq}, \\
    \label{eq:CDinequality}
    \textrm{Ineq. Cons.:} \quad  & \bm b^\mathrm{kd}_\mathrm{lb} \leq g^\mathrm{kd}_\mathrm{ineq}(\mathbf X^\text{kd}_k) \leq \bm b^\mathrm{kd}_\mathrm{ub}.
\end{alignat}
\end{subequations}
where the objective function (\ref{eq:CDMPCcost}) aims to track a predefined spatial momentum trajectory, state variable trajectory, and minimize ground reaction wrenches. The dynamics $f^{\text{kd}}$ in (\ref{eq:CDDynamics}) is the discrete-time dynamics of (\ref{eq:cd}-\ref{eq:hdot}). Additional constraints (\ref{eq:CDequality}-\ref{eq:CDinequality}) are similar to those of whole-body MPC.

\begin{figure}[!t]
\vspace{0.2cm}
    \center
    \includegraphics[clip, trim=1.2cm 0.5cm 1.2cm 0.3cm, width=1\columnwidth]{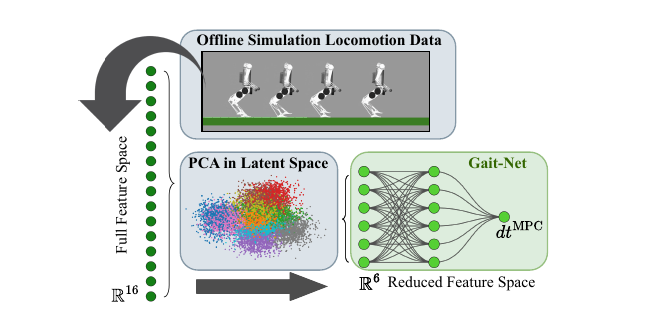}
    \caption{{\bfseries Illustration of the Gait-frequency Network}}
    \label{fig:gaitnet}
    \vspace{-0.5cm}
\end{figure}

It is worth noting that the kinematics aspect of the kino-dynamics MPC is computationally burdensome due to the direct optimization of full joint states. However, it may be embedded in the design of the MPC. As shown in \cite{dantec2024centroidal}, a CD-MPC chooses the spatial momentum vector and float-based states as the MPC states. This approach requires a somewhat accurate approximation of the evolution of the joint angles within the prediction horizon to construct meaningful spatial momentum trajectories. Hence,
\assumption{\label{assumption1}\textit{If the actual spatial momentum and pose evolution closely align with the designed trajectories, kinematic assurance can be implicitly embedded into the reference trajectory design rather than explicitly included in constraints. This enables an implicit kino-dynamic MPC approach.} }

\begin{figure*}[!t]
\vspace{0.2cm}
    \center
    \includegraphics[clip, trim=0.5cm 11.5cm 0.5cm 11.5cm, width=2\columnwidth]{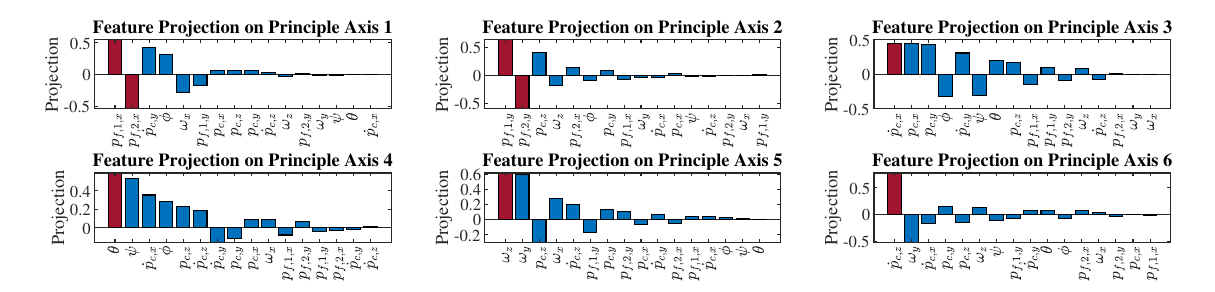}
    \caption{{\bfseries Feature Projection Bar Graphs along 6 Principle Axes.} The feature with the highest projection in each axis (red bar) is selected to be part of the new feature space. Note that along principle axes 1 and 2, both left and right legs are equally weighted with opposite signs, making the single Gait-Net suitable for both legs' prediction. }
    \label{fig:PCA}
    \vspace{-0.2cm}
\end{figure*}

In contrast, in BiConMP \cite{meduri2023biconmp}, the kinematic feasibility is guaranteed by a subsequent kinematics optimization after the MPC finds a feasible spatial momentum trajectory. In our work, we design and update these trajectories within each sequential CMPC subproblem and actively update the spatial reference based on the newly updated foot locations and float-based states within sequential iterations. We will discuss this key method in Sec. \ref{subsec:gaitnetmpc}. 

\section{Proposed Approach}
\label{sec:approach}

In this section, we introduce the proposed approaches in this work, including the Gait-frequency Network, the Gait-Net-augmented 
kino-dynamics MPC formulation, and the sequential MPC solving mechanism.

\subsection{Gait-frequency Network}
\label{subsec:gaitnet}

Instead of relying on heuristics or solving an additional optimization problem to determine step frequency from the desired foot location, we propose a lightweight Gait-frequency Network that maps current state feedback and desired foot placement to the preferred step duration for the upcoming stride, seamlessly integrating with the MPC framework while co-optimizing the target variables. An illustration of the Gait-Net is shown in Fig.  \ref{fig:gaitnet}. 

\subsubsection{Data collection}
We begin by using a whole-body MPC as our baseline controller to collect variable-frequency walking data. Besides a more accurate representation of kinematics and dynamics of the robot model, this approach is chosen for two key reasons: first, whole-body MPC outperforms other simplified-model-based control methods under strong unknown disturbances \cite{dantec2024centroidal}. Second, we utilize a MATLAB-based high-fidelity simulation framework, bypassing real-time hardware constraints. This enables the use of a more precise and robust interior-point-method-based nonlinear programming solver at higher control frequencies. The primary objective of this stage is to gather high-resolution data using a robust control algorithm under controlled disturbances in simulation.

We ran 15 different sets of walking simulations that have a total period of 600 seconds of walking data with a small-size humanoid model in the simulation. In each set, we command the robot to walk at a constant velocity in the range of $[0,\:1]$ \unit{m/s}. At the beginning of each stride, we generate a randomized step duration between $[150,\:400]$ \unit{ms}. In the first half of the walking simulation, the robot is commanded to walk without any disturbances, while in the second half, we apply randomized external impulses to the CoM of the robot every 2 seconds with the magnitude of $[10,\:100]$ \unit{N} for a duration of 0.2 seconds. 

Despite the relatively small training dataset, the Gait-Net and MPC combination effectively handles various scenarios with predicted step durations, as demonstrated in Sec. \ref{sec:Results}.

\subsubsection{Latent space feature reduction}
The collected data initially includes the robot's floating-base state variables and world-frame foot locations for both legs ($\mathbb R^{16}$). However, not all features/states are equally deterministic in the CoM space to determine the output of the network. To streamline the feature space for more efficient inference in each iteration while maintaining accurate predictions of variable MPC sampling time, we apply Principal Component Analysis (PCA) to reduce the input features. Specifically, we select one feature with a high absolute loading from each of the top six principal axes, as these features are minimally correlated and capture the most significant variance in the feature space. Fig.  \ref{fig:PCA} illustrates the PCA loadings of all features across six principal axes.

\subsection{Gait-Net-augmented Kino-dynamic MPC}
\label{subsec:gaitnetmpc}

\subsubsection{Motivation}
Given a fixed periodic contact sequence in MPC, one can vary the duration of each swing phase by altering the MPC sampling time $dt$ for the entire swing duration of each footstep made up of $h'$ MPC time steps. Hence the swing time $\Delta t = h'dt$. To achieve a variable-frequency walking MPC, we need to optimize the contact wrenches, contact locations, and MPC $dt$ all together.
In the discrete-time CD at time step $k$, 
\begin{equation}
\begin{aligned}
\label{eq:cd_discrete}
    \begin{bmatrix}
        \bm l_{G,k+1} \\ \bm k_{G,k+1}
    \end{bmatrix}
     = \begin{bmatrix}
        \bm l_{G,k} \\ \bm k_{G,k}
    \end{bmatrix} +
    \left[\begin{array}{c} 
    \sum_{i = 0}^{n_i}\bm f_i + m \bm g \\
    \sum_{i = 0}^{n_i}(\bm p_{f,i}-\bm p_c) \times \bm f_i + \bm \tau_i
    \end{array} \right]dt_k.
\end{aligned}
\end{equation}
Nonlinearity arises from the bilinear and multi-linear terms formed by the (cross) products of the three optimization variables.

In this section, we introduce a novel NN-augmented solving mechanism inspired by the popular SQP approach, allowing the MPC $dt$ to be concurrently determined by the Gait-Net alongside the optimization of other variables. 

\subsubsection{Implicit kinematics assurance in trajectory reference}
By selecting spatial momenta $\bm h$ and their primitive $\bm H$ (centroidal pose) as optimization variables, CD-MPC avoids including joint angles in the optimization process. However, generating these spatial reference trajectories still requires a corresponding whole-body joint space trajectory, which is a significant part of ensuring the kinematics feasibility of the optimization results. 
\begin{align}
    \bm h^\text{ref}_k = \bm A_G(\mathbf q^\text{ref}_k)\dot{\mathbf q}^\text{ref}_k,
\end{align}
\begin{equation}
\begin{aligned}
        \bm H^\text{ref}_k
        \approx \bm A_G(\mathbf q^\text{ref}_k)\mathbf q^\text{ref}_k - \sum^{k-1}_{i = 0} \dot{\bm A}_G(\mathbf q^\text{ref}_i,\dot{\mathbf q}^\text{ref}_i)\mathbf q^\text{ref}_i\:dt.
\end{aligned}
\end{equation}

Additionally, after obtaining the local foot location and CoM position trajectory solutions in each iteration, we perform a fast analytical inverse kinematics (IK) $f_\text{IK}$ to compute the corresponding joint trajectories for spatial momenta updates. The complete analytical IK and the approximation of $\bm H$ are provided in the Appendix. 

\remark{The spatial momentum and centroidal pose reference trajectories are updated in each sequential iteration to align with foot position updates, implicitly enforcing kinematic consistency in the reference trajectory. }

\subsubsection{Convex MPC Subproblem}
\label{subsubsec:cmpc}
To address the weak correlation between foot location and swing duration in the CD formulation, we integrate the proposed Gait-Net (Sec. \ref{subsec:gaitnet}) to predict the MPC sampling time at the beginning of each step based on the local foot and CoM location solutions at each sequential iteration. This process continues until the convergence condition is met. This also translates $dt$ from the optimization variable space to the parameter space for computation efficiency. 

\begin{figure}[!t]
\vspace{0.2cm}
    \center
    \includegraphics[clip, trim=4.5cm 12cm 4.1cm 12cm, width=1\columnwidth]{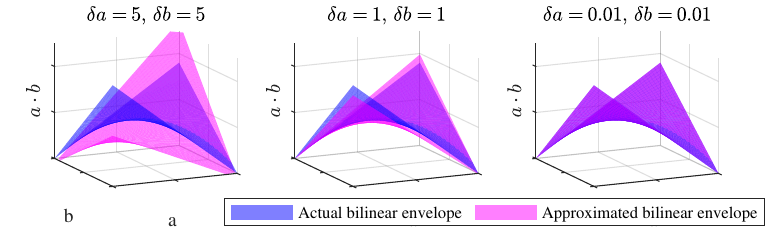}
    \caption{{\bfseries Bilinear Envelope Approximation by Neglecting Search Direction Product $\delta a\cdot\delta b$.}}
    \label{fig:bisurf}
    \vspace{-0.2cm}
\end{figure}

To linearize the bilinearly constrained dynamics constraint (\ref{eq:cd_discrete}) in the kino-dynamics MPC problem. We take inspiration from the SQP solving mechanism and solve the search direction $\delta$ of the bilinear variables, 
\begin{align}
    \bm f_i = \bm f^j_i +  \delta \bm f_i, \:\: \bm \tau_i = \bm \tau^j_i +  \delta \bm \tau_i, \nonumber\\
    \bm p_{f,i} = \bm p^j_{f,i} +\delta \bm p_{f,i}, \:\: \bm p_{c} = \bm p^j_{c} +\delta \bm p_{c},
\end{align}
where the $j$ superscript denotes the total solution from the previous sequential iteration $j$. 
The dynamics evolution of the angular momentum can  be simplified with the assumption:
\begin{assumption}
    \textit{With reasonable warm-start/initialization of the bilinear variables $a_0,\: b_0$, the bilinear product of the search directions are minimal and negligible as the sequential iteration $j$ increases, $\delta a^j \cdot \delta b^j \approx 0$, illustrated in Fig.  \ref{fig:bisurf}.} 
    \label{assump2}
\end{assumption}

Therefore, at iteration $j+1$, 
\begin{align}
\label{eq:lG}
    \bm l_{G,k+1} & = \bm l_{G,k} + m \bm g + \sum_{i = 0}^{n_i}\bm f^j_i dt_k +  \delta \bm f_i dt_k 
\end{align}
\begin{equation}
\allowdisplaybreaks
\begin{aligned}
    \label{eq:kG}
    \bm k_{G,k+1} & = \bm k_{G,k} + \sum_{i = 0}^{n_i}(\bm p^j_{f,i} - \bm p^j_c+\delta \bm p_{f,i}-\delta \bm p_c) \\ & \quad 
    \times (\bm f^j_i +  \delta \bm f_i)dt_k  + (\bm \tau^j_i + \delta \bm \tau_i)dt_k \\
    & = \bm k_{G,k} + \sum_{i = 0}^{n_i} \delta \bm \tau_i dt_k  -\bm f^j_i \times(\delta \bm p_{f,i}-\delta \bm p_c) dt_k \\ & \quad  
    +(\bm p^j_{f,i} - \bm p^j_c)\times \delta \bm f_i dt_k + 
    \underbrace{(\delta \bm p_{f,i}-\delta \bm p_c) \times \delta \bm f_i dt_k}_{\approx 0} \\ & \quad
     + \underbrace{(\bm p^j_{f,i} - \bm p^j_c)\times \bm f^j_i dt_k + \bm \tau^j_i dt_k}_\text{const.} 
\end{aligned}
\end{equation}
\remark{The linearization w.r.t. search directions $\delta$ and with \textit{Assumption} \ref{assump2} arrives mathematically identical to 1st-order Taylor expansion of bilinear constraints with the SQP algorithm \cite{boggs1995sequential}. }

\paragraph{Optimization Variables}
We choose the state variable to be the spatial momentum vector $\bm h$ and centroidal pose $\bm H$, and the control input variables to be the search directions of ground contact wrench, foot location, and MPC sampling time over a finite horizon $h$,
\begin{align}
    \bm {x} = \bigl\{ \bm H_{k};\: {\bm h}_{k}\bigr\}^{h}_{k=0}, \quad\quad\quad\quad\quad\\
    \bm {u} = \bigl\{\{\delta \bm f_{i,k},\: \delta \bm \tau_{i,k},\: \delta \bm p_{f,i,k}\}^{n_i}_{i=0},\:\delta \bm p_{c} \bigr\}^{h-1}_{k=0},
\end{align}

\paragraph{Objective Function}
The linearized finite horizon optimization problem can be formulated as 
\begin{alignat}{3}
    \label{eq:CDMPCcost2}
    \underset{\bm x^j, \bm {u}^j}{\text{min}} \: & \sum_{k = 0}^{h-1} \Bigg\| 
    \begin{bmatrix}
        \bm h_{k} \\ \bm H_{k} \\ \bm p^{j-1}_{f,k} +\delta \bm p_{f,k} \\ \bm p^{j-1}_{c,k} +\delta \bm p_{c,k}
    \end{bmatrix} -
    \begin{bmatrix}
        \bm h_{k}^{\text{ref}} \\ \bm H_{k}^{\text{ref}} \\ \bm p_{f,k}^{\text{ref}}\\ \bm p_{c,k}^{\text{ref}}
    \end{bmatrix}
    \Bigg\|^2 _{\bm L_1} \\ \nonumber & \quad +
    \Bigg\| \begin{bmatrix} 
    \bm f^{j-1}_i + \delta \bm f_i \\ \bm \tau^{j-1}_i  + \delta \bm \tau_i 
    \end{bmatrix} \Bigg\|^2 _{\bm L_2}
\end{alignat}
The objectives are to track spatial trajectories, foot reference trajectory, and CoM position trajectory while minimizing ground reaction wrenches. Note that the foot reference is constructed based on a preferred foot location through Gait-Net with a nominal MPC $dt$. This trajectory will be updated with each sequential iteration $j$ until convergence.
\paragraph{Linearized dynamics}
With \textit{Assumption} \ref{assump2} and search direction linearization in equation (\ref{eq:lG}-\ref{eq:kG}), we obtain the CD as a linear form with respect to our choice of control input and state variables. At time step $k$, the discrete dynamics is 
\begin{align}
\label{eq:cdlinear}
    \bm x_{k+1} = \mathbf A_k \bm x_{k} + \mathbf B_k \bm u_k + \mathbf C_k,
\end{align}
where $\mathbf A_k$ and $\mathbf B_k$ are the state space matrices. $\mathbf C_k$ contains only the constant terms shown in equation (\ref{eq:lG}-\ref{eq:kG}) in terms of optimization results obtained from the last sequential iteration $j-1$. By including a constant $1$ at the end of the state variables, $\bm x'_{k} = [\bm x_{k};\:1]$, we can then obtain the discrete dynamics in a linearized state-space form, as similarly outlined in \cite{di2018dynamic,chen2024learning}.

\paragraph{Linear momentum and CoM position}
The dynamics evolution of CoM position $\bm p_c$ is embedded in the linear momentum equation as an equality constraint:
\begin{align}
    \bm p^{j-1}_{c,k+1} +\delta \bm p_{c,k+1}  = \bm p^{j-1}_{c,k} +\delta \bm p_{c,k} + \frac{\bm l_{G,k}}{m}\:dt_k.
\end{align}

\paragraph{Friction pyramid}
The friction pyramid is a conservative approximation of the friction cone when the pyramid is inscribed, such that the pyramid friction coefficient is then approximated as 
\begin{align}
    \mu_{\Box} = \frac{\sqrt{2}}{2}\mu,
\end{align}
where $\mu$ is the actual ground friction coefficient. The linear inequality constraint is
\begin{align}
    \nonumber 
    -\mu_{\Box}  ({f}_{i,z}^{j-1}+\delta f_{i,z}) \leq  \big\{({f}_{i,x}^{j-1}+\delta f_{i,x}), 
    \: ({f}_{i,y}^{j-1}+\delta f_{i,y})\bigr\} \\
    \leq \mu_{\Box}  ({f}_{i,x}^{j-1}+\delta f_{i,x}) \quad \quad
    \label{eq:frictionCons}
\end{align}

\paragraph{Contact-switching Constraints}
For periodic walking gait, we use a binary contact schedule $\bm \sigma \in \mathbb R^{n_i\times h}$ to describe the contact switches for each contact point $i$. Hence, the ground reaction wrench can be switched on or off based on whether a leg is in swing or stance, 
\begin{align}
    \label{eq:contactConstraint1}
     \sigma_{i,k}
    \begin{bmatrix}
        \bm f_\text{min} \\ \bm \tau_\text{min} 
    \end{bmatrix} \leq
    \begin{bmatrix}
        \bm f^{j-1}_i + \delta \bm f_i \\ \bm \tau^{j-1}_i  + \delta \bm \tau_i
    \end{bmatrix} \leq 
    \sigma_{i,k}
    \begin{bmatrix}
        \bm f_\text{max} \\ \bm \tau_\text{max} 
    \end{bmatrix}  
\end{align}
In addition to the control input saturation, we also enforce stationary foot location for each footstep while the contact schedule $\sigma_{i,k} = 1$ for contact point $i$.

\begin{figure*}[!t]
\vspace{0.2cm}
		\center
		\includegraphics[clip, trim=0cm 12cm 0.2cm 0cm, width=2\columnwidth]{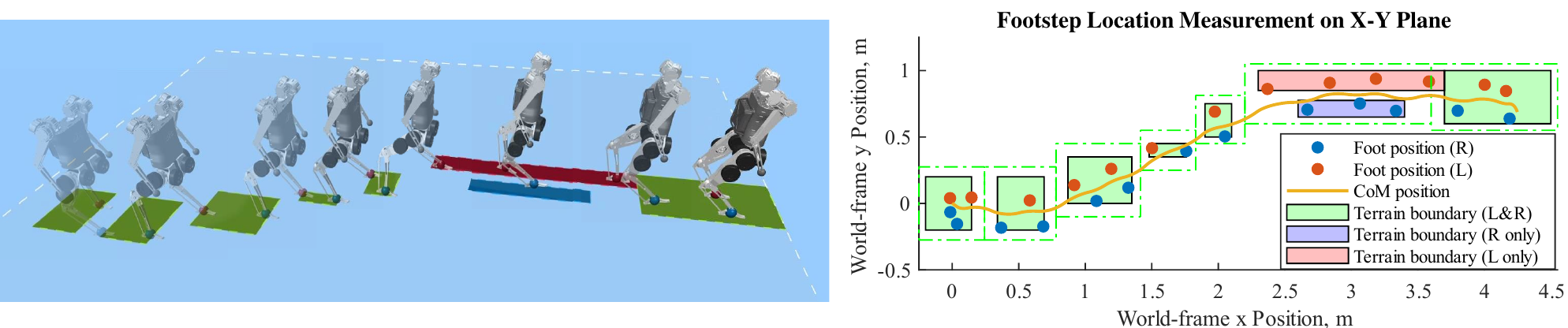}
		\caption{{\bfseries Locomotion over 3-D Stepping-stone Terrain.} Simulation snapshots (left) and plot of measured foot locations (right). In the plot, only foot locations that are on the ground are visualized. The green dashed-line bounding box represents the CoM position threshold that triggers the foot location constraints for the corresponding stepping stone patch.}
		\label{fig:footlocation}
		\vspace{-0.2cm}
\end{figure*}

\paragraph{Dynamic Range of Foot location}
With the future foot location as part of the optimization variables, the constraints can be directly applied with only one-step preview data in discrete terrain - the bounds of the foot location are dependent on the position of the robot,  
\begin{align}
    \bm p_{f,\text{min}}(\bm p_c) \leq \bm p^{j-1}_{f,k} +\delta \bm p_{f,k} \leq \bm p_{f,\text{max}}(\bm p_c)
    \label{eq:KDfoot}
\end{align}
A visualization of the constraint-triggering conditions is illustrated in Fig. \ref{fig:footlocation}. 

\begin{algorithm}[h!]
\caption{Gait-Net-augmented Sequential CMPC}
\label{alg:gaitMPC}
\begin{algorithmic}[1]
\Require $\mathbf q, \: \dot{\mathbf q}, \: \mathbf q^\text{cmd}, \: \dot{\mathbf q}^\text{cmd}$
\State \textbf{intialize} $\bm x_0 = f_\text{j2m}(\mathbf q, \: \dot{\mathbf q}), \: \bm u^0 =\bm u_\text{IG}, \: dt^0 = 0.05$ 
\State $\{ \mathbf q^\text{ref},\:\dot{\mathbf q}^\text{ref},\:\bm p_f^\text{ref}\} = f_\text{ref} \big(\mathbf q, \: \dot{\mathbf q}, \: \mathbf q^\text{cmd}, \: \dot{\mathbf q}^\text{cmd} \big)$
\State $\bm x^\text{ref} = f_\text{j2m}(\mathbf q^\text{ref},\:\dot{\mathbf q}^\text{ref},\:\bm p_f^\text{ref})$
\State $ j = 0$ 
\While{$j \leq j_\text{max} \:\text{and}\: \bm \eta \leq \delta \bm u  $} 
\State $\delta \bm u^{j} = \texttt{cmpc}(\bm x^\text{ref},\:\bm p_f^\text{ref},\:\bm p_c^\text{ref},\: \bm x_0,\: dt^j, \: \bm u^j)$
\State $\bm u^{j+1} = \bm u^j + \delta \bm u^j$ 
\State $dt^{j+1} = \mathcal N_\text{GN}(\mathbf q, \: \dot{\mathbf q},\: \bm p_f^{j})$
\State $\{ \bm x^\text{ref},\:\bm p_f^\text{ref}\}= f_\text{IK}(\bm p_f^{j},\:\bm p_c^{j},\: dt^{j+1})$
\State $j=j+1$
\EndWhile \\
\Return $\bm u^{j+1} $
\end{algorithmic}
\end{algorithm}
\subsubsection{Gait-Net-augmented Sequential MPC Algorithm}
Algorithm \ref{alg:gaitMPC} outlines the procedure for solving the proposed kino-dynamic MPC with sequential CMPC subproblems and Gait-Net. 

In the initialization stage (1-4), $f_\text{j2m}$ describes the mapping from joint-space general-coordinate states to spatial momenta. $f_\text{ref}$ construct a reference trajectory in generalized coordinate with a nominal sampling duration $dt^0$. Within the sequential CMPC iterations (5-11), the iteration is terminated until reached max iteration $j_\text{max}$ or the search direction reaches the desired tolerance $\bm \eta$. In each iteration, the CMPC subproblem described in \ref{subsubsec:cmpc} is solved via QP; the MPC sampling time is updated through Gait-Net policy $\mathcal N_\text{GN}$. Subsequently, the reference trajectories are updated to reflect the latest foot location and MPC $dt$. 

\begin{figure}[!t]
\vspace{0.2cm}
     \centering
     \begin{subfigure}[b]{0.5\textwidth}
         \centering
	   \includegraphics[clip, trim=0cm 10.4cm 7.4cm 0cm, width=1\columnwidth]{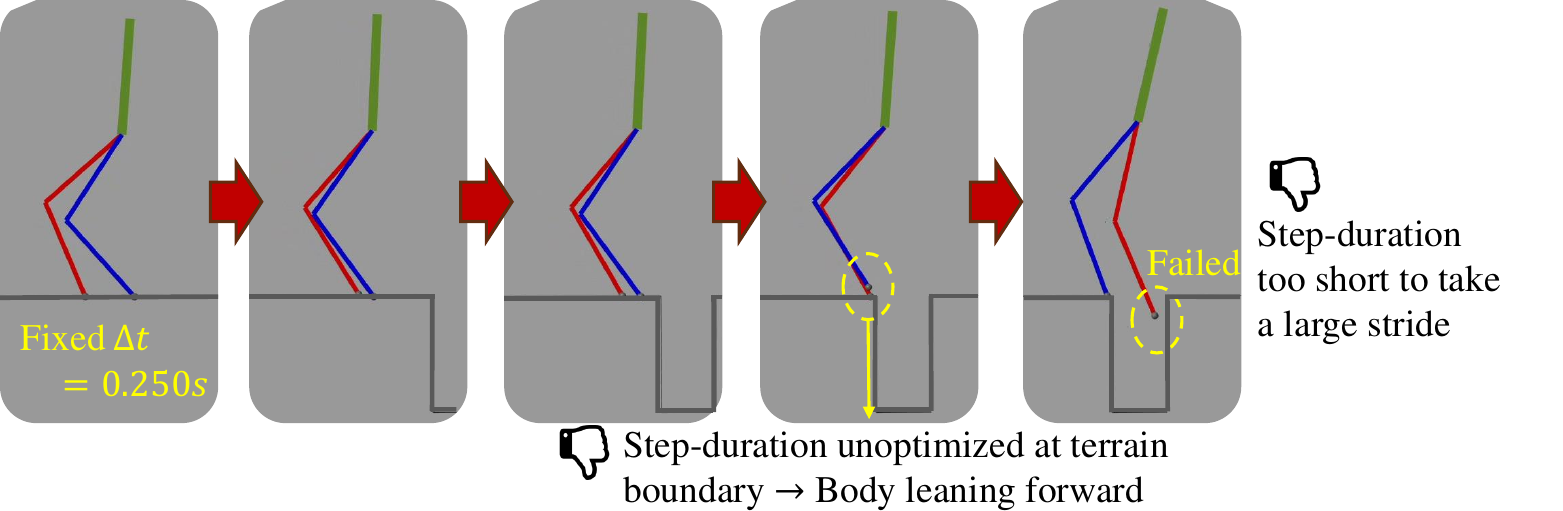}
          \caption{Baseline 1: Fixed step duration for every step.}
          \vspace{0.2cm}
          \label{fig:comp1_3}
     \end{subfigure}
     \begin{subfigure}[b]{0.5\textwidth}
        \includegraphics[clip, trim=0cm 11cm 7.4cm 0cm, width=1\columnwidth]{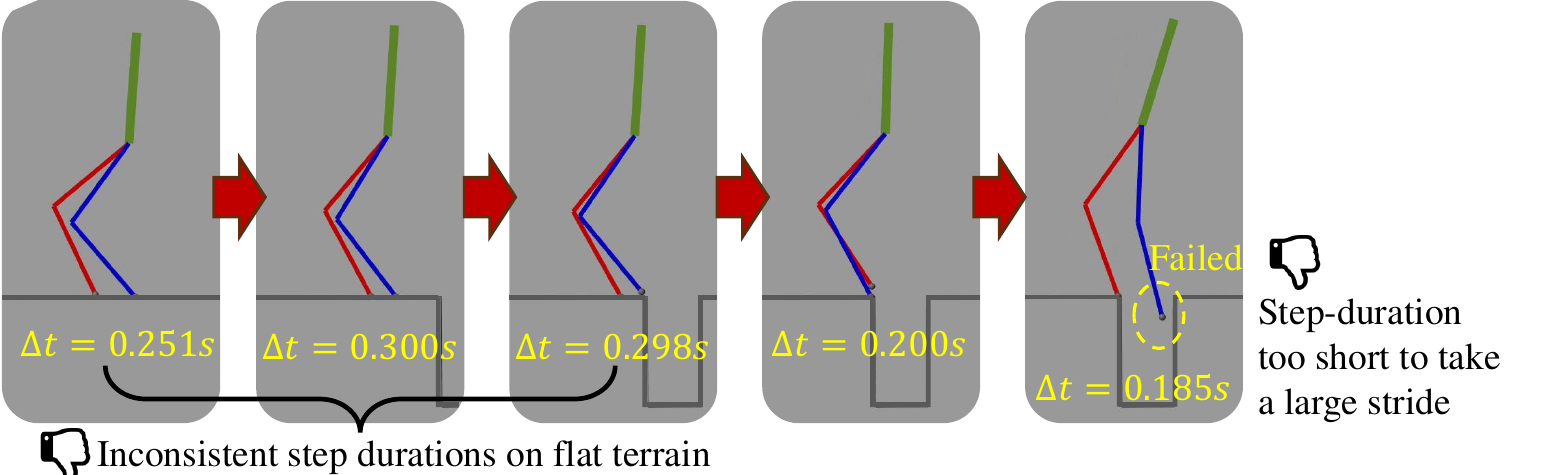}
	\caption{Baseline 2: Solving step duration as part of optimization variables in NMPC.}
        \vspace{0.2cm}
	\label{fig:comp1_1}
     \end{subfigure}
     \begin{subfigure}[b]{0.5\textwidth}
         \centering
	   \includegraphics[clip, trim=0cm 10cm 7.4cm 0cm, width=1\columnwidth]{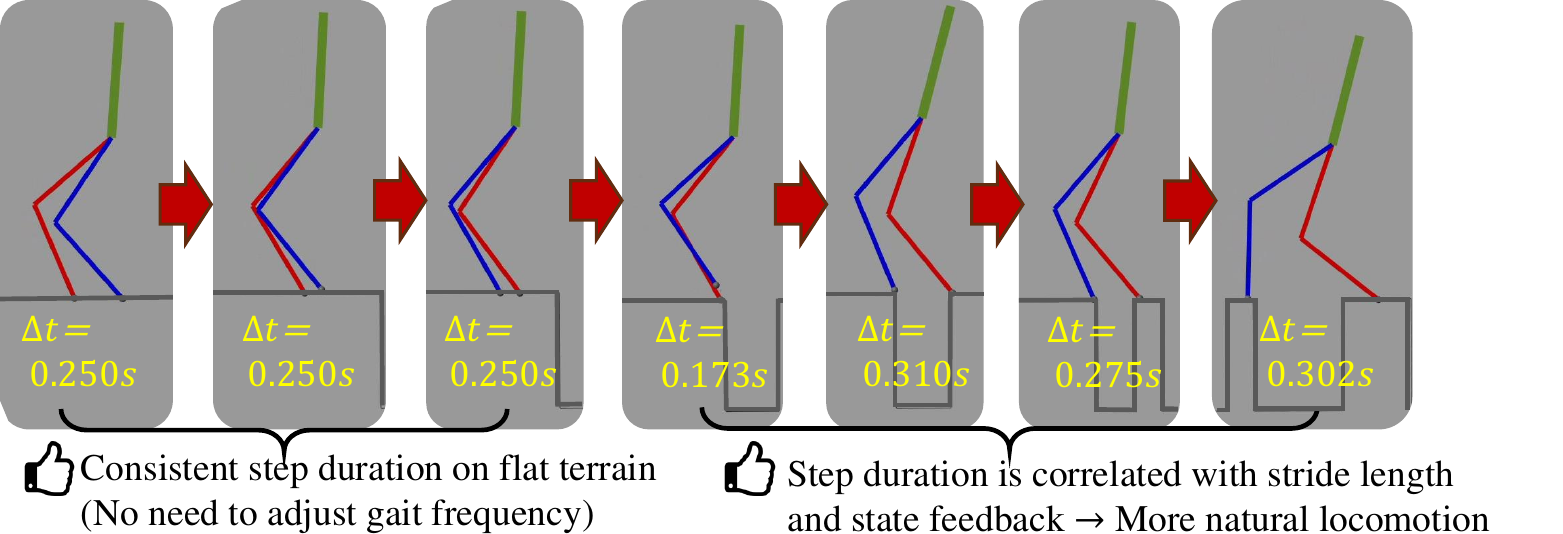}
          \caption{Proposed: Gait-Net-augmented Kino-dynamic MPC.}
          \vspace{0.4cm}
          \label{fig:comp1_2}
     \end{subfigure}
     \begin{subfigure}[b]{0.48\textwidth}
         \centering
	   \includegraphics[clip, trim=0cm 3cm 0cm 0cm, width=1\columnwidth]{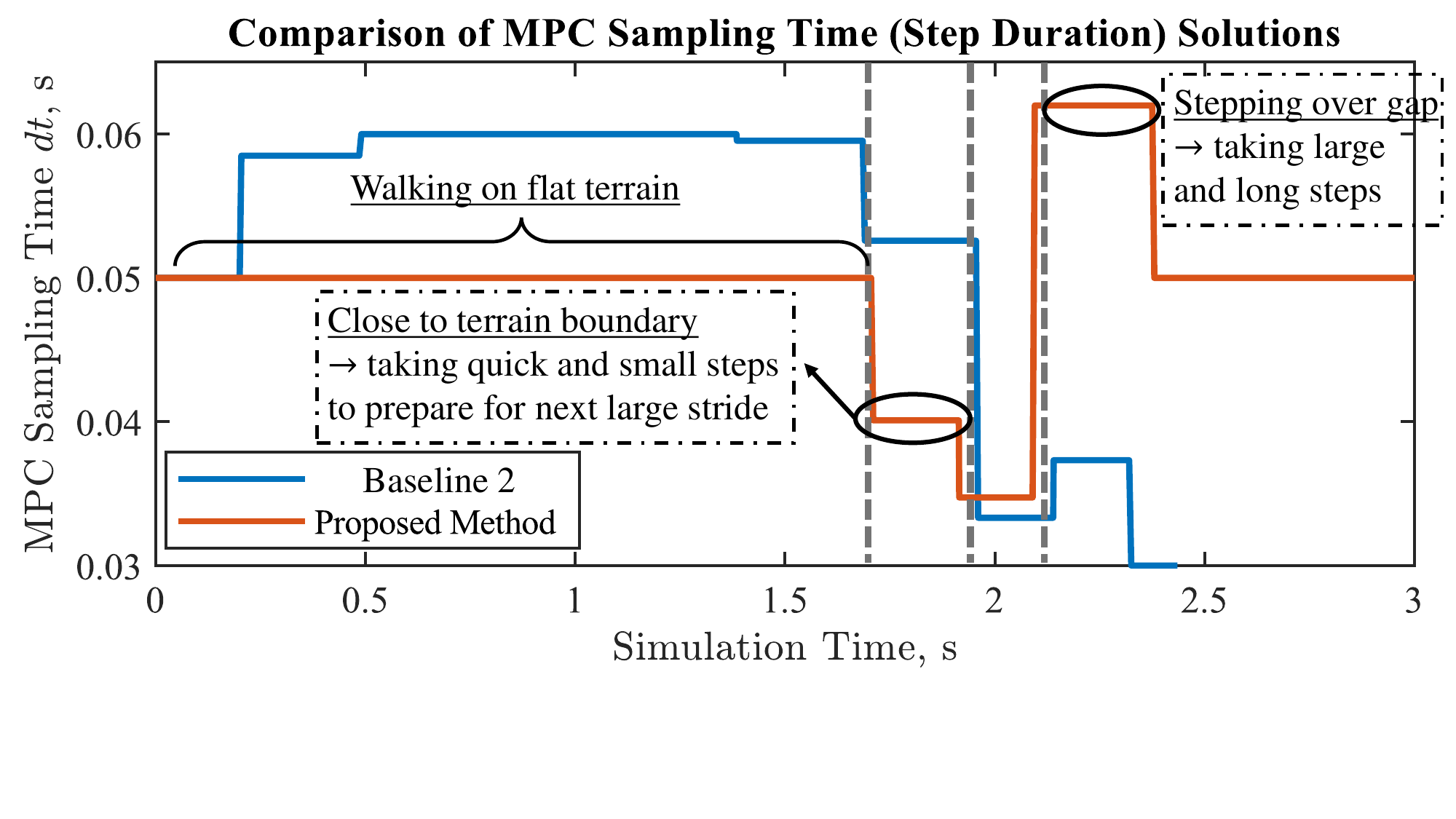}
          \caption{Comparison of MPC $dt$ (interpreted as step duration).}
          \label{fig:dt_comp}
     \end{subfigure}
     \caption{{\bfseries{Comparison of Discrete Terrain Locomotion Performance in 2D Simulation.}} }
        \label{fig:comp1}
\end{figure}

\begin{figure*}[!t]
\vspace{-0.1cm}
		\center
		\includegraphics[clip, trim=0cm 11.3cm 0.2cm 0cm, width=1.9\columnwidth]{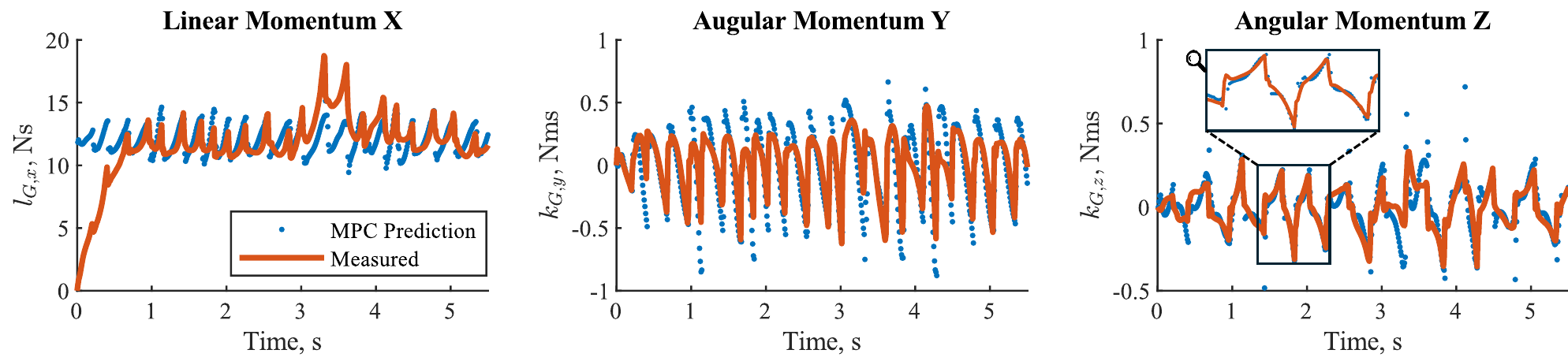}
		\caption{{\bfseries Spatial Momenta Measurement vs. MPC Prediction along $l_{G,x},\:k_{G,y},\:$and $k_{G,z}$ of 3-D Stepping-stone Simulation Results.}}
		\label{fig:h_tracking}
		\vspace{-0.2cm}
\end{figure*}

\remark{As the sequential iteration progresses, the reference trajectories $\{ \bm x^\text{ref},\:\bm p_f^\text{ref}\}$ are continuously updated to match closely to the real spatial momentum and pose trajectories based on the latest kinematics results. This process inherently embeds an \textit{implicit kinematics assurance} within the framework.}

\remark{The Gait-Net-augmented kino-dynamic MPC is run at the beginning of each footstep to determine a local step duration in terms of MPC $dt$. The rest of this footstep will incorporate the same $dt$ without the inference of Gait-Net and solve only the contact location and wrenches.}

\section{Results}
\label{sec:Results}

In this section, we present the main results of the proposed control framework in both high-fidelity simulation and on small-size humanoid hardware. 

\subsection{Validation Setup}
We perform the simulation validations through a custom MATLAB/Simulink-based simulator with Simscap Multi-body Library. The robot model has two variants: the 2-D 5-link bipedal robot consists of 7 degree-of-freedom (DoF), and the HECTOR V2 humanoid robot \cite{li2023hector} with 24 DoF, 5 actuated joints on each leg, and 4 actuated joints on each arm. Both robots are miniature-sized and have a total leg length of $44 \unit{cm}$. 

The simulations are performed on a Ubuntu 20.04 machine with an AMD Ryzen 7900X CPU. The optimization problems are solved through CasADi \cite{Andersson2019} MATLAB interface with \texttt{sqpmethod} solver for SQP problems, and \texttt{qpoases} solver \cite{Ferreau2014} for QP subproblems. 

The humanoid hardware equips Intel NUC i5 mini-PC, which solves the sequential CMPCs in real-time with the \texttt{qpoases} solver in C++. The Gait-Net is lightweight and its inference is done with CPU computation.

The state estimation of the robot relies on onboard sensors. The CoM translational states are estimated by a linear Kalman filter fusing accelerometer data and robot kinematics, with correction by Intel T265 camera.
Orientation is estimated using a complementary filter with IMU gyro data and T265 correction. 
Note that in this work, the entire terrain data is known to the robot. However, to mimic how real-time perception acquires close-by static terrain data, in our experiments, only a \textit{one-step preview} of next-step terrain boundary is available at each step, interpreted as a function of CoM position (eqn. (24)).

\subsection{Numerical Analysis of Gait-Net}
We first compare the performance of the trained Gait-Net with (1) full feature space provided by simulation data, meaning the input space consists of the full CoM states (position $\bm p_c$, velocity $\dot{\bm p}_c$, Euler angles $\bm \Theta$, and angular velocity $\bm \omega$) and desired foot locations, totaling $\mathbb R^{16}$; (2) reduced feature space after performing PCA and selecting top features from 6 principle axes, shown in Fig. \ref{fig:PCA}. Table \ref{tab:NetsolveTime} compares the normalized inference time of the Gait-Net between the two methods and their RMSE in predicting the MPC sampling time $dt$ with test sets. 

\begin{table}[h]
\setlength\extrarowheight{2pt}
\setlength{\tabcolsep}{3.5pt}
\vspace{0cm}
    \centering
    \caption{Gait-Net Performance Comparisons}
    \begin{tabular}{ | m{12em} | m{8em} | m{8em} | }
    \hline \hline
	 \makecell[c]{Feature Space}  & \makecell[c]{Full (baseline)} & \makecell[c]{\textbf{PCA Selection}}  \\
    \hline \hline
    Feature Size & \makecell[c]{$\mathbb {R}^{16}$}  & \makecell[c]{$\mathbb {R}^{6}$} \\ \hline
    Normalized Inference Time  & \makecell[c]{\underline{1.00}} & \makecell[c]{\textbf{0.74}} \\ \hline
    Prediction RMSE & \makecell[c]{$4.98\mathrm{e}{-3}$} & \makecell[c]{$6.79\mathrm{e}{-3}$}   \\
    \hline \hline
    \end{tabular}
    \label{tab:NetsolveTime}
    \vspace{-0.2cm}
\end{table}

We also provide a numerical analysis to validate the current Gait-Net’s robustness to sensor noise, trained with noise-free data. Table \ref{tab:1} compares MPC $dt$ predictions on a standardized test set run and repeated under varying IMU and encoder noise levels, simulated in MATLAB using Allan variance from both IMU factory specs and measured IMU noise parameter data. Despite these noises, the worst-case RMSE differs by only 0.39 ms from the noise-free case, demonstrating Gait-Net's reliability under realistic sensor conditions.
\begin{table}[h]
\setlength\extrarowheight{2pt}
\setlength{\tabcolsep}{3.5pt}
    \centering
    \caption{Gait-Net Performance w/ Sensor Noises}
    \begin{scriptsize}
    \begin{tabular}{ | m{18em} | m{14em} | }
    \hline \hline
	 \makecell[c]{Sensor Noise Setup}  & \makecell[c]{RMSE w.r.t. Ground Truth}   \\
    \hline \hline
    Noise-free (currently employed) & \makecell[c]{$\mathbf{6.79\mathrm{e}{-3}}$}  \\ \hline
    W/ factory IMU noise parameters  & \makecell[c]{$6.95\mathrm{e}{-3}$}  \\ \hline
    W/ measured IMU noise parameters & \makecell[c]{$7.16\mathrm{e}{-3}$}   \\ \hline
    W/ measured IMU and joint encoder noise & \makecell[c]{$\mathbf{7.18\mathrm{e}{-3}}$} \\
    \hline \hline
    \end{tabular}
    \end{scriptsize}
    \label{tab:1}
    \vspace{-0.2cm}
\end{table}

\subsection{Comparison of Kino-dynamic MPC Solving Methods}
Next, we present a comparison of our proposed kino-dynamic MPC and its solving method along with a few other state-of-the-art methods, shown in Table \ref{tab:mpcsolveTime}.

\begin{table}[t]
\vspace{0.2cm}
\centering
\setlength\extrarowheight{1pt}
\setlength{\tabcolsep}{2.5pt}
\caption{Comparison with Solving Approaches (3-D)}
\label{tab:mpcsolveTime}
\begin{scriptsize}
  \begin{tabular}{ | m{6.5em} | m{3.5em}  | m{3.5em} | m{4.5em} | m{4.5em} | m{4.25em} | } 
  \hline
  \hline
  \makecell[c]{Method} & \multicolumn{3}{ c |}{SQP} & \makecell[c]{AD(3)} & \makecell[c]{\bf{Proposed}} \\ 
  \hline
  \hline
  MPC Model & \makecell[c]{WBD} &  \makecell[c]{Explicit\\KD} & \multicolumn{2}{ c |}{CD}& \makecell[c]{Implicit\\KD} \\
  \hline
  Optimization Variable Size & \makecell[c]{$\mathbb {R}^{78h}$} & \makecell[c]  {$\mathbb {R}^{60h}$} & \makecell[c]{$\mathbb {R}^{34h-22}$} & \makecell[c]{$\mathbb {R}^{27h-15}$ \\ $\mathbb {R}^{21h-9}$ \\$\mathbb {R}^{16h-4}$} &  \makecell[c]{$\mathbb {R}^{33h-21}$}\\ 
  \hline
  Normalized Solve Time & \makecell[c]{12.05} & \makecell[c]{7.28} & \makecell[c]{\underline{1.00}} & \makecell[c]{2.58} &  \makecell[c]{\textbf{0.76}}\\ 
  \hline
   Average Number of QPs & \makecell[c]{18.2} & \makecell[c]{17.7} & \makecell[c]{19.5} & \makecell[c]{27.5} &  \makecell[c]{\textbf{17.1}}\\ 
  \hline
  \hline
\end{tabular}
\end{scriptsize}
\vspace{0.1cm}
\begin{flushleft}
\footnotesize{\scriptsize{ \quad WBD - \textit{Whole-body dynamics}; \: CD - \textit{Centroidal dynamics}; \: KD - \textit{Kino-dynamics}.}}    
\end{flushleft}
\vspace{-0.2cm}
\end{table}

\begin{figure}[!t]
\vspace{0.2cm}
    \center
    \includegraphics[clip, trim=4.5cm 9cm 4.8cm 9cm, width=1\columnwidth]{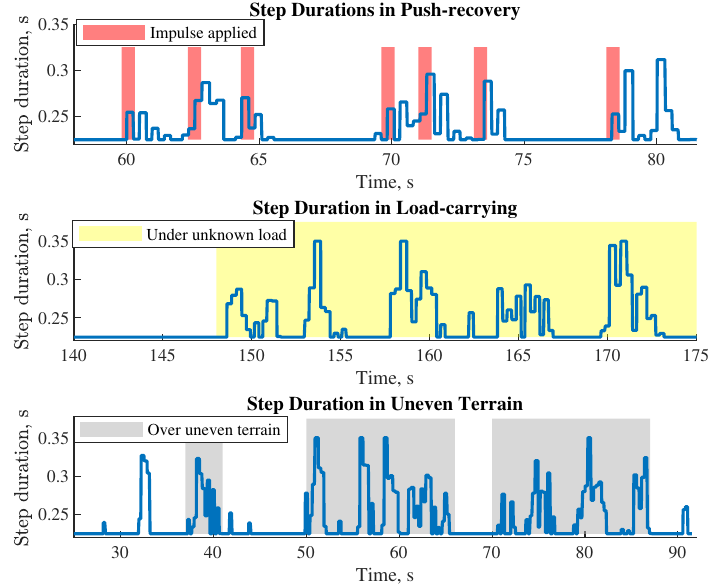}
    \caption{{\bfseries Variable Step Durations under Unknown Disturbances.} Hardware experiment of (1) Push-recovery, Fig. \ref{fig:title}(a); (2) Carrying unknown load, Fig. \ref{fig:title}(b); (3) Locomotion over uneven terrain, Fig. \ref{fig:title}(c).}
    \label{fig:disturbance}
    \vspace{-0.2cm}
\end{figure}

 The compared whole-body MPC (Sec. \ref{subsec:wbmpc}) and explicit kino-dynamic MPC (Section. \ref{subsec:cdmpc}) consider MPC sampling time as additional optimization variables.
 The average solve time of each solving method is normalized by the baseline (CD-MPC solved with SQP). The average number of QPs refers to the QP (sequential) subproblems performed until convergence. These values are measured by running 1000 time steps of each solving mechanism. The alternating direction (AD) approach follows a similar algorithm in \cite{meduri2023biconmp} but with 3 alternating directions (\textit{a.k.a}., mountain-climbing method \cite{konno1976cutting}). All problems are solved until the same termination threshold is reached in this comparison.

Whole-body MPC is expected to be the most time-consuming variant when solved to full convergence, this aligns with the motivation for adopting a lighter kino-dynamic model for more efficient deployment while maintaining a reasonable trade-off in model fidelity. In CD-MPC, the AD approach requires numerous iterations to solve the 3 separate subproblems, resulting in a longer overall solve time. In contrast, our proposed method is more efficiently designed, achieving a faster convergence rate compared to the baseline approaches.

Then, we validate the significance of Gait-Net to determine the step duration (interpreted as MPC $dt$) within the sequential solving mechanism, rather than solving it as a part of the optimization. We compare locomotion performance in 2-D numerical simulation with a 5-link bipedal robot and discrete terrain gaps in the range of $[5,\;15]$ cm. Fig. \ref{fig:comp1} compares snapshots of our proposed kino-dynamic MPC with (1) fixed gait frequency, (2) variable gait frequency by solving step duration as an additional optimization variable, and (3) variable gait frequency by the proposed Gait-Net. With fixed step duration, the periodic gait is un-optimized especially near the discrete terrain boundary. Solving step duration in optimization is also not optimal, as the MPC $dt$ does not have a strong physical correlation with the step location, resulting in random/inconsistent solutions that are unintuitive to interpret in motion. With the proposed method, the step duration is highly dependent on the robot state and next foot location through Gait-Net, rendering more natural and whole-body-kinematic-aware walking behavior.

\begin{figure}[!t]
\vspace{0.2cm}
     \centering
     \begin{subfigure}[b]{0.5\textwidth}
         \centering
	   \includegraphics[clip, trim=0cm 7cm 8.7cm 0cm, width=1\columnwidth]{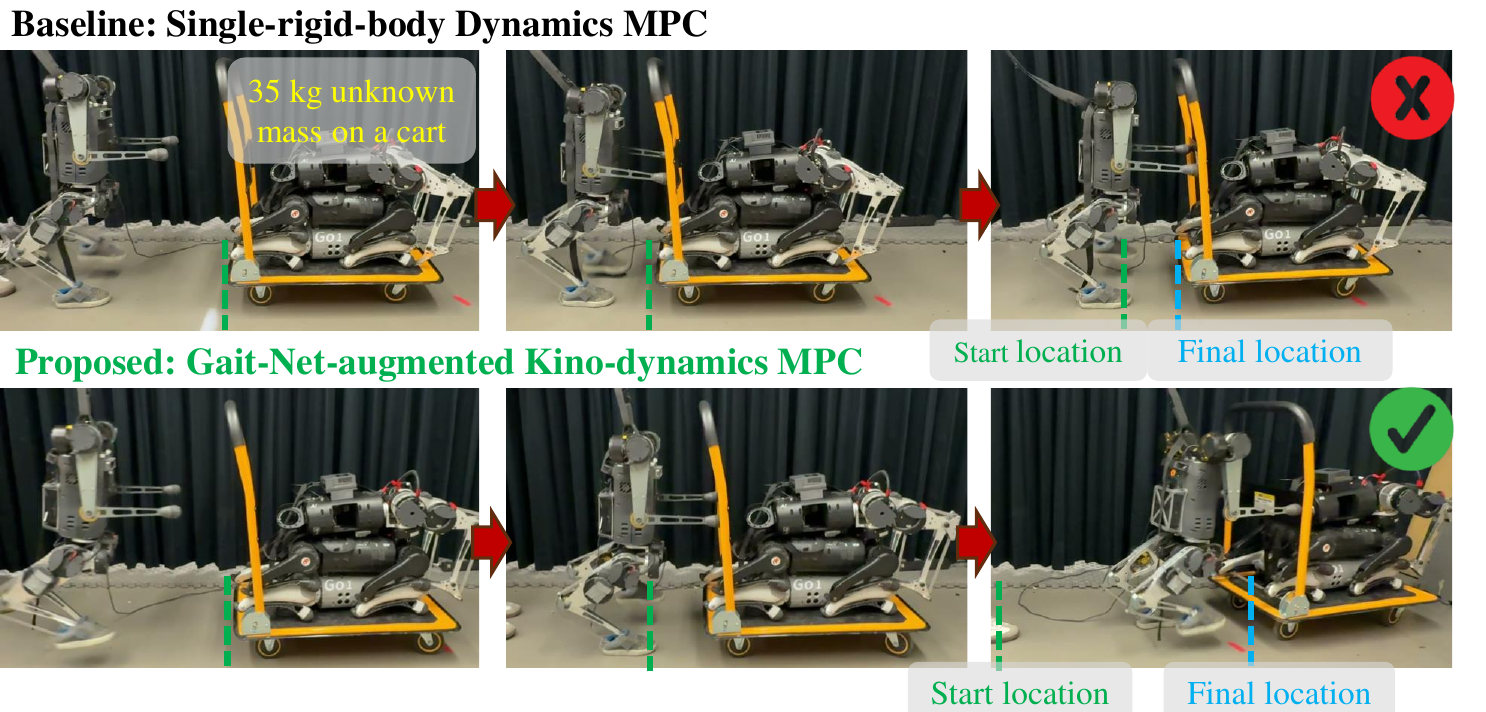}
          \caption{Simulation Snapshots. Green and blue dashed lines represent the start and final location of the cart.}
          \vspace{0.2cm}
          \label{fig:comppush1}
     \end{subfigure}
     \begin{subfigure}[b]{0.475\textwidth}
        \includegraphics[clip, trim=0cm 4.5cm 19cm 9cm, width=1\columnwidth]{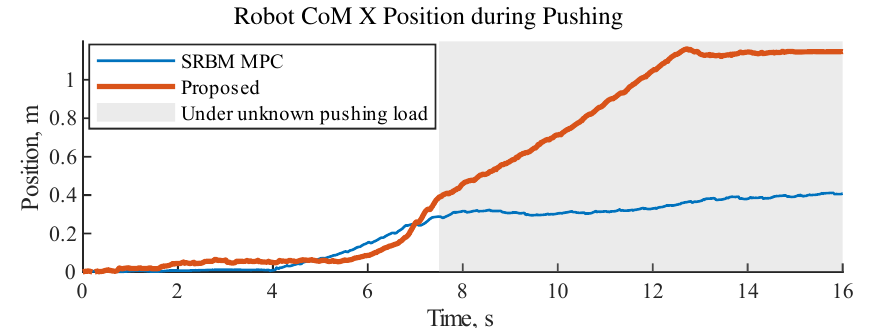}
	\caption{CoM X position comparison.}
        \vspace{0.2cm}
	\label{fig:comppush2}
     \end{subfigure}
     \begin{subfigure}[b]{0.475\textwidth}
        \includegraphics[clip, trim=0cm 10.5cm 19cm 2.5cm, width=1\columnwidth]{figures/loco_manipulation.pdf}
	\caption{Foot location vs. MPC $dt$ prediction in loco-manipulation.}
        \vspace{0.2cm}
	\label{fig:comppush3}
     \end{subfigure}
     \caption{{\bfseries Pushing Carts with Unknown Payload.} The time instances of the robot hand's contact with the cart are synchronized in the comparison plot to enhance visualization.}
    \label{fig:compare_pushing}
    \vspace{-0.2cm}
\end{figure}

\subsection{3-D Locomotion over Discrete Terrain}
\label{subsec:discrete_terrain}

We showcase the capability of the proposed method in the 3-D environment with a line-foot humanoid robot. In simulation, we construct a stepping-stone course for the robot to navigate through with the proposed method. Given a forward velocity command, $\dot {\bm p}^\text{ref}_{c,x} = 0.75$ m/s, the control goal is to reach the end of the course while avoiding stepping outside of the stepping stone patches, illustrated in Fig. \ref{fig:footlocation}. Note that the green patches allow both feet to step on, the red patch and the blue patch only allow the corresponding colored foot to step on. The center locations of the feet are plotted in Fig. \ref{fig:footlocation}, demonstrating the capability of our variable-frequency walking method in traversing challenging terrains dynamically and the satisfaction of foot location constraints with the proposed method.

Associated plots of spatial momenta are also provided in Fig. \ref{fig:h_tracking}, where the scatter data represent the reference trajectories from MPC at the beginning of each footstep for better visualization purposes. The plots demonstrate a good prediction of spatial momenta evolution with the proposed Gait-Net-augmented sequential method. 


\subsection{Hardware Experiment Validation}
We conduct hardware experiments on a 3-D small-size humanoid robot with 5-DoF legs. The hardware experiment snapshots are presented in Fig. \ref{fig:title}. The readers are encouraged to watch the supplementary video for better visual aids. 

\begin{figure}[!t]
\vspace{0.2cm}
    \center
    \includegraphics[clip, trim=0cm 2cm 0cm 0cm, width=0.95\columnwidth]{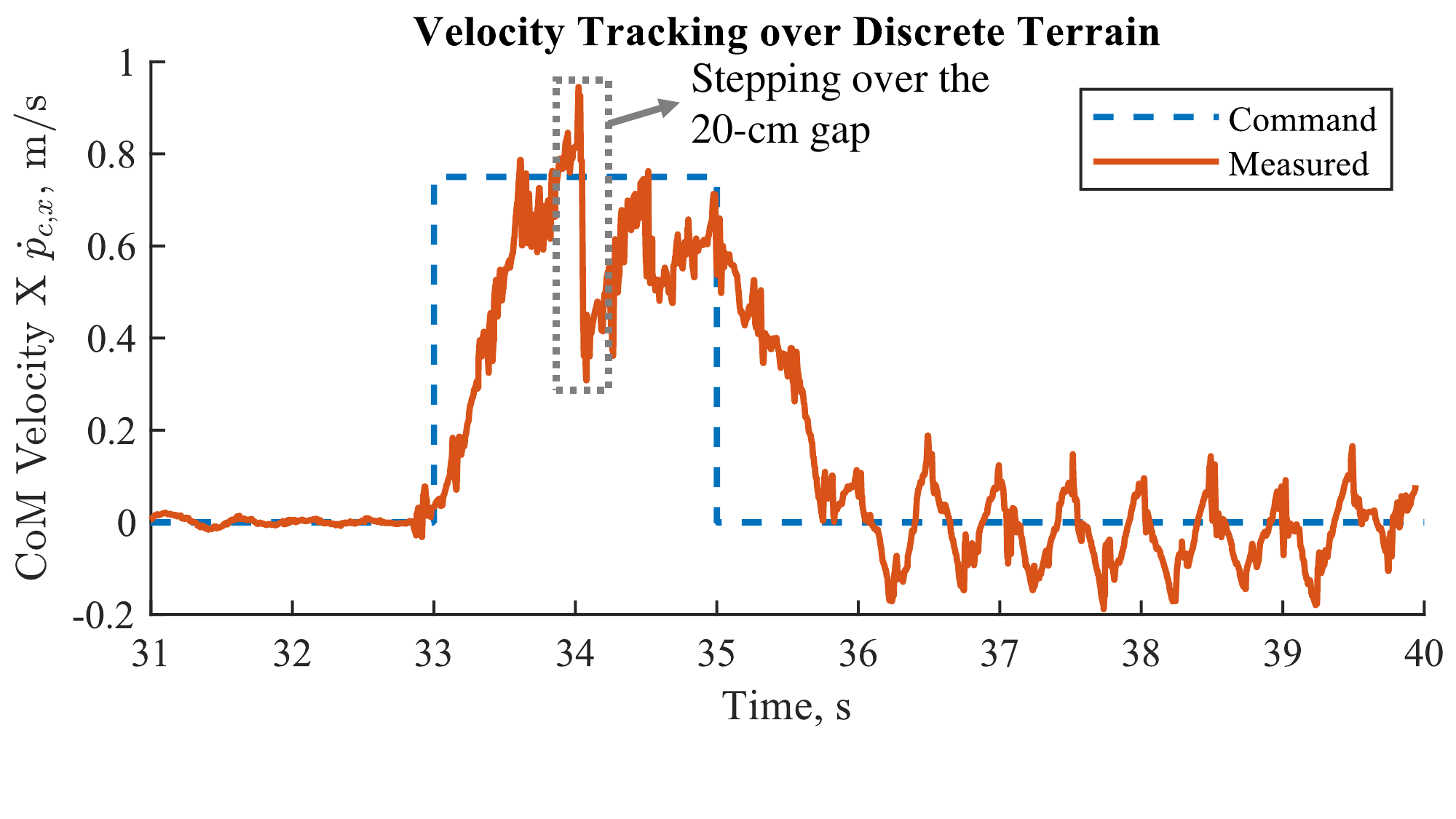}
    \caption{{\bfseries Velocity X Tracking Plot.} In discrete terrain locomotion hardware experiment with a 20-cm gap. }
    \label{fig:vx}
    \vspace{-0.2cm}
\end{figure}

\subsubsection{Locomotion under Unknown Perturbations}
First, we demonstrate the robustness of the proposed control framework in handling unknown perturbations, including push-recovery, handling unknown payload, and uneven terrain locomotion, as illustrated in Fig. \ref{fig:title}(a-c). The controller dynamically adjusts step duration at each step to maintain balance while handling these unknown disturbances. Fig. \ref{fig:disturbance} presents the variable step durations from Gait-Net-augmented MPC. We clip the MPC $dt$ predictions to be within the bound of $[0.045,\:0.07]$ s to ensure hardware feasibility. On relatively flat terrain with minimal perturbations, the MPC optimizes $dt$ to the nominal value of 0.045 s, equivalent to a 0.225 s step duration. While under disturbances, the step durations are adjusted. Notably, foot positions are not the primary factors in predicting the MPC step duration; instead, the Gait-Net is trained to incorporate additional deterministic state features for improved adaptation.

\subsubsection{Baseline Comparison in Loco-manipulation}

In this comparison, the robot is tasked with negotiating and pushing a cart of a 35-kg unknown load (219$\%$ of the robot's mass) using two approaches: (1) SRBM MPC with a fixed gait frequency and (2) the proposed Gait-Net-augmented kino-dynamic MPC. Snapshots of the experiment and CoM position comparisons are presented in Fig. \ref{fig:compare_pushing}.
The results show that the SRBM approach struggles to handle such a large unknown load, whereas the proposed method successfully manages the disturbance.
In Fig. \ref{fig:comppush3}, it is observed that over 10 forward-pushing trials, longer step durations favor near foot locations for more effective pushing actions, while shorter step durations help recover balance with larger steps.
It is worth noting that the observed performance difference is not solely due to gait frequency adaptation but rather benefits from the integration of the proposed techniques.

\subsubsection{Discrete Terrain Locomotion}

Following the validation objectives outlined in Sec. \ref{subsec:discrete_terrain}, we conduct dynamic locomotion experiments on discrete terrains with hardware. While the robot has prior knowledge of the terrain map, terrain discontinuity information is provided only as one-step preview.

We first present a locomotion experiment on flat ground featuring a 20-cm-wide virtual gap, marked by red tape, with a commanded velocity of 0.75 \unit{m/s}. Snapshots of this experiment are shown in Fig. \ref{fig:title}(c), and the velocity tracking performance is shown in Fig. \ref{fig:vx}. It is important to note that foot placement is defined by the center of the foot, allowing a small portion of the toe or heel to step into the red-taped area. This is feasible because the CWC constraint adapts to ensure compliance with the line-foot constraint along the edges, as long as the center of the foot is within the safe area. 

Next, we increase the complexity of the discrete terrain course by introducing a 10-cm-wide 3-cm-high obstacle, and a 15-cm-wide virtual gap, marked by red tape. However, the obstacle's height is unknown to the robot. Snapshots of this experiment are presented in Fig. \ref{fig:title}(d).
An interesting observation is that while traversing the obstacle, the robot deliberately leans to its right, allowing the right foot to step on flat ground while the left leg is optimized for a slightly longer step duration to clear the obstacle successfully.

\subsubsection{Analysis on NN-augmented SQP Sensitivity}

While SQP is robust, integrating a neural network into a sequential solver does not guarantee convergence.. Therefore, we monitor the search direction solutions. If a non-negative slope appears, we revert to nominal $dt$ and SQP for guaranteed convergence. Within the conducted experiments, only 2.8$\%$ instances required this fallback, adding 23$\%$ average solve time for such cases.

\begin{figure}[t!]
  \centering
  \includegraphics[clip, trim=4.5cm 11.25cm 5cm 11cm, width=0.95\columnwidth]{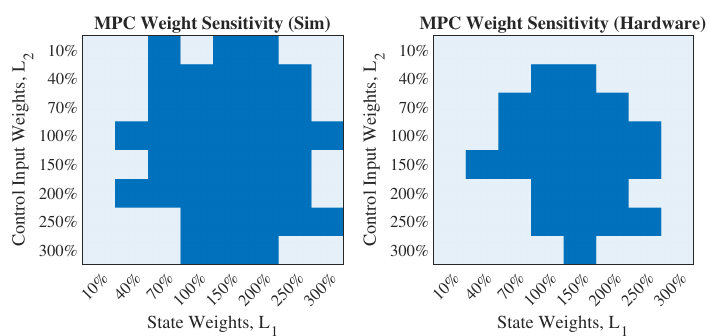}
  \caption{MPC weight parameter sensitivity: robot performance with 10$\%$ to 300$\%$ of original hand-tuned values. (Dark blue: stable parameter region) \label{fig:sens}}
  \vspace{-0.6cm}
\end{figure}

Our approach also offers flexibility in tuning and selecting optimal control parameters. While the current set of parameters, listed in the Appendix, is manually tuned through iterative testing, we found that the range of suitable MPC weight values is relatively broad. Fig. \ref{fig:sens} presents a sensitivity analysis of the weight parameters $\bm L_1$ and $\bm L_2$, where their values are varied from 10$\%$ to 300$\%$ relative to the original hand-tuned values, both in simulation and on hardware. The results indicate that the controller exhibits a reasonable level of robustness to parameter variations.

\section{Limitation}
\label{sec:futureWork}


Currently, the direct foot position constraint relies on real-time CoM position feedback, which eliminates the need for if-else conditions within the optimization process. However, this approach limits consideration to only the immediate next step, as the one-step preview data is applicable solely to the upcoming step. As a result, constraints for subsequent steps within the prediction horizon must be intentionally deactivated, ensuring that only the next step location is constrained using the available preview terrain data.

\section{Conclusions}
\label{sec:Conclusion}
In conclusion, we propose a Gait-Net-augmented kino-dynamic MPC framework for controlling humanoid robot locomotion over discrete terrains, enabling variable-frequency walking behaviors while explicitly constraining foot locations. Our approach addresses the challenge of concurrently optimizing contact location, contact force, and step duration in periodic walking motions through a solving mechanism that relies solely on sequential CMPCs. To achieve this, we leverage a pre-trained Gait-Net to bridge the gap between contact location and step duration within a centroidal dynamics formulation. This integration allows the spatial momentum and pose trajectories to be iteratively updated during the solving process, providing implicit kinematic assurance through trajectory reference design. The simulation and hardware experiments demonstrated effective control performance for 3-D stepping stone locomotion in a highly dynamic manner. The proposed control framework allows superior perturbation-resistant locomotion capabilities over baseline approaches. We also successfully achieved walking on a small-sized humanoid robot at a speed of 0.75 m/s while safely avoiding undesired terrain gaps and obstacles, even in the presence of terrain height perturbations.

\section{Acknowledgement}
\label{sec:Acknowledgement}

This work is supported by the USC Departmental Startup Fund.

The authors would like to extend special thanks to Lokesh Krishna for his valuable insights on the deployment engineering of Gait-Net in numerical optimization; and Dr. Jingyi Liu for the discussion of math details.

\balance
\bibliographystyle{IEEEtranN}

\bibliography{reference.bib}

\newpage
\appendices
\section{Approximation of Spatial Momentum Primitives (Centroidal Pose)}
\label{sec:appendix}

The primitive of the spatial momentum vector (centroidal pose) in terms of generalized coordinate states $\mathbf q, \dot{\mathbf q}$ can expressed as an indefinite integral of $\bm h$:
\begin{equation}
\label{eq:app1}
    \bm H = \int \bm h\:dt= \int \bm A_G\dot{\mathbf q}\: dt 
\end{equation}

We apply integration by parts, 
\begin{equation}
    \int u\:dv = uv - \int v\: du,
\end{equation}

And set,
\begin{equation}
    u = \bm A_G,\quad dv = \dot{\mathbf q}\: dt.
\end{equation}

Then,
\begin{equation}
    du = \dot{\bm A}_G\:dt,\quad v = \int \dot{\mathbf q}\: dt = \mathbf q.
\end{equation}

Hence, substituting into equation (\ref{eq:app1},
\begin{equation}
\begin{aligned}
        \bm H &
        =\bm A_G\mathbf q - \int \biggl( \int \dot{\mathbf q} dt \biggl)\dot{\bm A}_G\:dt + \text{const.}
\end{aligned}
\end{equation}

We apply the Riemann Sum Approximation over a finite number of discrete time steps $k$ from $t_0$ to $t_k$ with step size $\Delta t$:
\begin{equation}
\begin{aligned}
        \int^{t_k}_{t_0}\bm h\: dt \approx
        \bm A_G\mathbf q - \sum^{k-1}_{i = 0} \dot{\bm A}_G\mathbf q_i\:\Delta t
\end{aligned}
\end{equation}
\vspace{0.5cm}

\section{Analytic IK Solution of 5-DoF Legs with Line-foot}
\begin{figure}[h!]
    \includegraphics[clip, trim=-5cm 5.2cm 13cm 0.3cm, width=1\columnwidth]{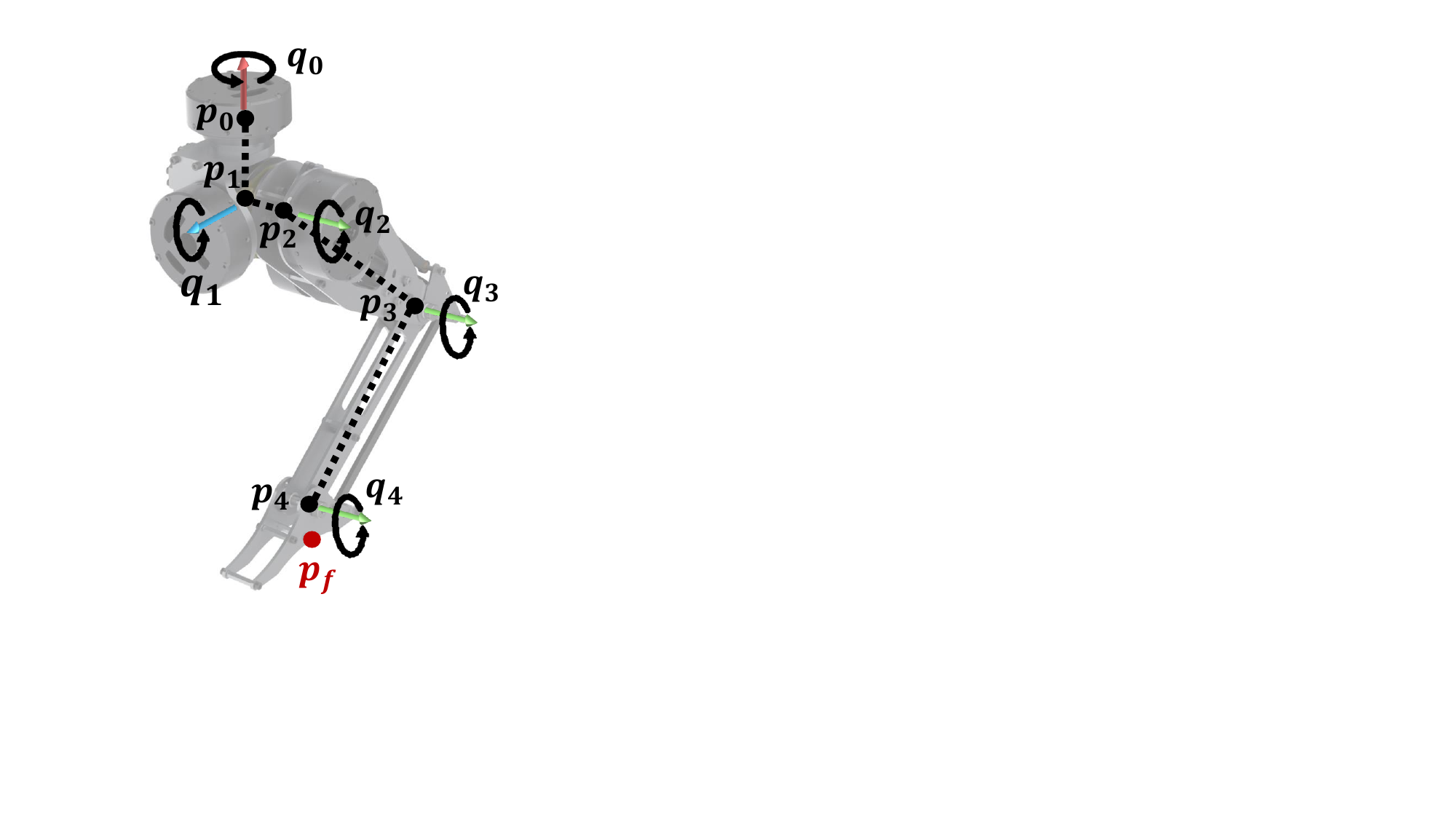}
    \caption{Kinematic Definition of 5-DoF Leg with Line-foot.}
    \label{fig:leg}
\end{figure}

In this appendix, we present the detailed analytical solution derivation of the 5-DoF leg inverse kinematics (IK) for fast computation. Fig. \ref{fig:leg} shows the kinematic tree. The default zero-joint-angle configuration is when the leg is straight (no knee-bend) with the foot perpendicular to the leg.

To efficiently map end-effector position $\bm p_f \in \mathbb R^3$ to the corresponding feasible joint positions $\bm q \in \mathbb R^5$, we use additional orientation constraints during the computation of the IK, such that the foot pitch and yaw angles are always zeros w.r.t. the body, $R_{f,y} = 0$, and $R_{f,z} = 0$. Due to the kinematic design of this leg, these additional orientation constraints can be mapped directly to yaw joint $q_0$ and ankle pitch joint $q_4$, such that

\begin{equation}
    q_0^\text{des} = 0,
\end{equation}
\begin{equation}
    q_4^\text{des} = - q_2^\text{des}- q_3^\text{des}-\theta.
\end{equation}

The rest of the joint position $\{q_1,\:q_2,\:q_3\}$ can be mapped to $\bm p_f^\text{des}$ with kinematic geometry, where,

\begin{equation}
    q_1^\text{des} = \arcsin({\frac{r^\text{des}_z}{r^\text{des}_{1,yz}}}) + \arcsin({\frac{s_j\:r_{21,y}}{r^\text{des}_{1,yz}}}),
\end{equation}
\begin{equation}
    q_2^\text{des} = \arccos({\frac{r^\text{des}_{1,xz}}{2l}})- s_j \cdot \arccos({\frac{r^\text{des}_{2,yz}}{r^\text{des}_{1,xz}}}),
\end{equation}
\begin{equation}
    q_3^\text{des} = 2\arcsin({\frac{r^\text{des}_{1,xz}}{2l}})- \pi,
\end{equation}
where the kinematic parameter $r_{21,y}$ is the y-direction distance between joint location $\bm p_2$ and $\bm p_1$. 
$\bm r^\text{des}_{1}$ is the leg frame distance vector from the desired foot position to the roll joint,
\begin{equation}
    \bm r^\text{des}_{1} = R^\intercal\bm p_f^\text{des}-\bm p_1,
\end{equation}
where $\bm p_1$ in body frame can be obtained from the CoM position and the distance vector between CoM and joint 1 $\bm r_{c1}$,
\begin{equation}
    \bm p_1 = R^\intercal(\bm p_c - \bm r_{c1}),
\end{equation}

$r^\text{des}_{1,yz}$ is the absolute distance of $\bm r^\text{des}_{1}$ along y-z plane,
\begin{equation}
    r^\text{des}_{1,yz} = \sqrt{(r^\text{des}_{1,y})^2+(r^\text{des}_{1,z})^2}
\end{equation}

$r^\text{des}_{2,yz}$ is the absolute distance from foot to thigh pitch joint,  
\begin{equation}
    r^\text{des}_{2,yz} = \sqrt{(r^\text{des}_{1,yz})^2+r_{21,y}^2}
\end{equation}

$r^\text{des}_{2,xz}$ is the absolute distance from foot to thigh pitch joint along x-z plane,
\begin{equation}
    r^\text{des}_{1,xz} = \sqrt{(r^\text{des}_{1,x})^2+(r^\text{des}_{1,y})^2+(r^\text{des}_{1,z})^2 - r_{21,y}^2}
\end{equation}
\vspace{0.2cm}

\section {Physical and Control Parameters}
In this Appendix, we provide all the physical parameters of the robots and the tuning parameters of the controllers used. 

\begin{table}[H]
	\vspace{-0.1cm}
	\centering
        \caption{2-D Bipedal Robot}
	\begin{tabular}{cccc}
		\hline
		Parameter & Symbol & Value & Units\\
		\hline
		Mass & $m$    & 10.0 & $\unit{kg}$  \\
		  Body MoI  & $\bm I_{yy}$  & 0.15 & $\unit{kg}\cdot \unit{m}^2$ \\
		Leg Lengths & $l_1$    & 0.22 & 
          $\unit{m}$  \\
          & $l_2$    & 0.22 & $\unit{m}$  \\
		\hline 
	\end{tabular}
	\vspace{0.2cm}
\end{table}	
\clearpage

\begin{table}[H]
	\vspace{0.2cm}
	\centering
        \caption{3-D Humanoid Robot}
	\begin{tabular}{cccc}
		\hline
		Parameter & Symbol & Value & Units\\
		\hline
		Mass & $m$    & 16.00 & $\unit{kg}$  \\
		Body MoI  & $\bm I_{xx}$  & 0.069 & $\unit{kg}\cdot \unit{m}^2$ \\
		& $\bm I_{yy}$ & 0.058  & $\unit{kg}\cdot \unit{m}^2$ \\
		& $\bm I_{zz}$ & 0.007  & $\unit{kg}\cdot \unit{m} ^2$ \\
		Leg Lengths & $l_1$    & 0.22 & 
          $\unit{m}$  \\
          & $l_2$    & 0.22 & $\unit{m}$  \\
		 Foot Length & $l_{f}$    & 0.15 & $\unit{m}$  \\
		\hline 
	\end{tabular}
    \vspace{0.2cm}
\end{table}

\begin{table}[H]
	\vspace{0.2cm}
	\centering
	\caption{2-D Gait-Net-Augmented  Kino-dynamic MPC Control Parameters (Sim)}
	\begin{tabular}{ccc}
	\hline
	Parameter & Symbol & Value \\
	\hline
	Prediction length & $h$  &  10 \\
        Swing phase & $h'$ & 5 \\
        Step size & $dt$  &  0.03 - 0.06 s \\
        Friction coef. & $\mu$  &  0.7 \\
        Max. force & $f_\text{max}$  &  250 N \\
        Min. force & $f_\text{min}$  &  10 N \\
        Max. Torque & $\tau_\text{max}$ & 33.5 Nm / 67 Nm (knee)\\
        Control weights & $\bm L_1^{\bm h}$  &  $[20, 20, 20]$ \\
        & $\bm L_1^{\bm H}$ & $[200, 200, 300]$ \\
        & $\bm L_1^{\bm {p}_f}$ & $[0.01, 0.01]$ \\
        & $\bm L_1^{\bm {p}_c}$ & $[100, 300]$ \\
        & $\bm L_2^{\bm f}$ & $[0.0001, 0.0001]$ \\
	\hline 
	\end{tabular}
	\vspace{0.2cm}
\end{table}	


\begin{table}[H]
	\vspace{0.2cm}
	\centering
	\caption{3-D Whole-body MPC Control Parameters (Sim)}
	\begin{tabular}{ccc}
	\hline
	Parameter & Symbol & Value \\
	\hline
	Prediction length & $h$  &  10 \\
        Swing phase & $h'$ & 5 \\
        Step size & $dt$  &  0.045 - 0.07 s \\
        Friction coef. & $\mu$  &  0.7 \\
        Max. force & $f_\text{max}$  &  500 N \\
        Min. force & $f_\text{min}$  &  10 N \\
        Max. Torque & $\tau_\text{max}$ & 33.5 Nm / 67 Nm (knee)\\
        Control weights & $\bm Q_1^{\bm q_\text{fb}}$  &  $[100, 100, 100, 150, 150, 250]$ \\
        & $\bm Q_1^{\bm q_{\text{j},i}}$  &  $[50,50,50,50,50]$ \\
        & $\bm Q_2^{i}$ & $[0.001, 0.001, 0.001, 0.001, 0.001]$ \\
        & $\bm Q_3^{\bm f}$ & $[0.0001, 0.0001, 0.0001]$ \\
        & $\bm Q_3^{\bm \tau}$ & $[0.001, 0.001, 0.001]$ \\
        & $\bm Q_4^{\dot{\bm q}_\text{fb}}$  &  $[1, 1, 1, 1, 1, 5]$ \\
        & $\bm Q_4^{\dot{\bm q}_{\text{j},i}}$  &  $[1,1,1,1,1]$ \\
	\hline 
	\end{tabular}
	\vspace{0.2cm}
\end{table}	

\begin{table}[H]
	\vspace{0.2cm}
	\centering
	\caption{3-D Explicit Kino-dynamic MPC Control Parameters (Sim)}
	\begin{tabular}{ccc}
	\hline
	Parameter & Symbol & Value \\
	\hline
	Prediction length & $h$  &  10 \\
        Swing phase & $h'$ & 5 \\
        Step size & $dt$  &  0.045 - 0.07 s \\
        Friction coef. & $\mu$  &  0.7 \\
        Max. force & $f_\text{max}$  &  500 N \\
        Min. force & $f_\text{min}$  &  10 N \\
        Max. Torque & $\tau_\text{max}$ & 33.5 Nm / 67 Nm (knee)\\
        Control weights & $\bm R_1$  &  $[10, 10, 10, 20, 20, 20]$ \\
        & $\bm R_2^{\bm q_\text{fb}}$  &  $[100, 100, 100, 150, 150, 250]$ \\
        & $\bm R_2^{\bm q_{\text{j},i}}$  &  $[50,50,50,50,50]$ \\
        & $\bm R_2^{\dot{\bm q}_\text{fb}}$  &  $[1, 1, 1, 1, 1, 5]$ \\
        & $\bm R_2^{\dot{\bm q}_{\text{j},i}}$  &  $[1,1,1,1,1]$ \\
        & $\bm R_2^{\bm f}$ & $[0.00001, 0.00001, 0.00001]$ \\
        & $\bm R_2^{\bm \tau}$ & $[0.0001, 0.0001, 0.0001]$  \\
	\hline 
	\end{tabular}
	\vspace{0.2cm}
\end{table}	

\begin{table}[H]
	\vspace{0.2cm}
	\centering
	\caption{3-D Gait-Net-Augmented Kino-dynamic MPC Control Parameters (Sim)}
	\begin{tabular}{ccc}
	\hline
	Parameter & Symbol & Value \\
	\hline
	Prediction length & $h$  &  10 \\
        Swing phase & $h'$ & 5 \\
        Step size & $dt$  &  0.045 - 0.07 s \\
        Friction coef. & $\mu$  &  0.7 \\
        Max. force & $f_\text{max}$  &  500 N \\
        Min. force & $f_\text{min}$  &  10 N \\
        Max. Torque & $\tau_\text{max}$ & 33.5 Nm / 67 Nm (knee)\\
        Control weights & $\bm L_1^{\bm h}$  &  $[10, 10, 10, 10, 10, 10]$ \\
        & $\bm L_1^{\bm H}$  &  $[200, 200, 200, 200, 200, 200]$ \\
        & $\bm L_1^{\bm p_{f,i}}$ & $[0.1, 0.1, 0.1]$ \\
        & $\bm L_1^{\bm p_{c}}$ & $[100, 100, 250]$ \\
        & $\bm L_2^{\bm f}$ & $[0.0001, 0.0001, 0.0001]$ \\
        & $\bm L_2^{\bm \tau}$ & $[0.001, 0.001, 0.001]$ \\
	\hline 
	\end{tabular}
	\vspace{0.2cm}
\end{table}	

\begin{table}[H]
	\vspace{0.2cm}
	\centering
	\caption{3-D Gait-Net-Augmented Kino-dynamic MPC Control Parameters (Hardware)}
	\begin{tabular}{ccc}
	\hline
	Parameter & Symbol & Value \\
	\hline
	Prediction length & $h$  &  10 \\
        Swing phase & $h'$ & 5 \\
        Step size & $dt$  &  0.045 - 0.07 s \\
        Friction coef. & $\mu$  &  0.7 \\
        Max. force & $f_\text{max}$  &  500 N \\
        Min. force & $f_\text{min}$  &  10 N \\
        Max. Torque & $\tau_\text{max}$ & 33.5 Nm / 67 Nm (knee)\\
        Control weights & $\bm L_1^{\bm h}$  &  $[5, 5, 5, 1, 1, 1]$ \\
        & $\bm L_1^{\bm H}$  &  $[400, 400, 400, 300, 300, 300]$ \\
        & $\bm L_1^{\bm p_{f,i}}$ & $[0.1, 0.1, 0.1]$ \\
        & $\bm L_1^{\bm p_{c}}$ & $[400, 400, 500]$ \\
        & $\bm L_2^{\bm f}$ & $[0.00001, 0.00001, 0.00001]$ \\
        & $\bm L_2^{\bm \tau}$ & $[0.0001, 0.0001, 0.0001]$ \\
	\hline 
	\end{tabular}
	\vspace{0.2cm}
\end{table}	

\begin{table}[H]
	\vspace{0.2cm}
	\centering
	\caption{Gait-Net-Augmented Sequential CMPC Solver Parameters}
	\begin{tabular}{ccc}
	\hline
	Parameter & Symbol & Value \\
	\hline
	Max. iterations & $j_\text{max}$  &  50 \\
        Position tolerance & $\eta^\text{pos}$ & 0.00001 \\
        Force tolerance & $\eta^{f}$  &  0.01 \\
        Moment tolerance & $\eta^\tau$  &  0.001 \\
	\hline 
	\end{tabular}
	\vspace{0.2cm}
\end{table}	

\end{document}